\documentclass[a4paper,12pt,twoside]{article}

\usepackage{hyperref}
\usepackage{graphicx}
\usepackage{amsmath, amssymb}
\usepackage{cite} 
\usepackage[top=1.5cm,bottom=1.5cm,left=2cm,right=2cm,includeheadfoot]{geometry}
\usepackage[nottoc]{tocbibind}

\newcommand{\doubleEnter}{\par\vspace{\baselineskip}}

\DeclareSymbolFont{greekletters}{OML}{cmr}{m}{it}
\DeclareMathSymbol{\varrho}{\mathalpha}{greekletters}{"25}

\begin{document}
\begin{titlepage}
  \center
  \textsc{\huge University of Groningen}\\[0.3cm]
  \textsc{\large Faculty of Science and Engineering}\\[0.5cm]
  \vfill
  \rule{\linewidth}{0.5mm} \\[0.4cm]
  { \huge \bfseries Controlling Recurrent Neural Networks by\\ \vspace{0.3cm}Diagonal Conceptors}\\[0.4cm] 
  \rule{\linewidth}{0.5mm} \\[0.4cm]
  \vspace{0.5cm}
  \textsc{\Large J.P. De Jong}\\[0.5cm]
  \vfill
  \vspace{2cm}
  \includegraphics[width=0.2\textwidth]{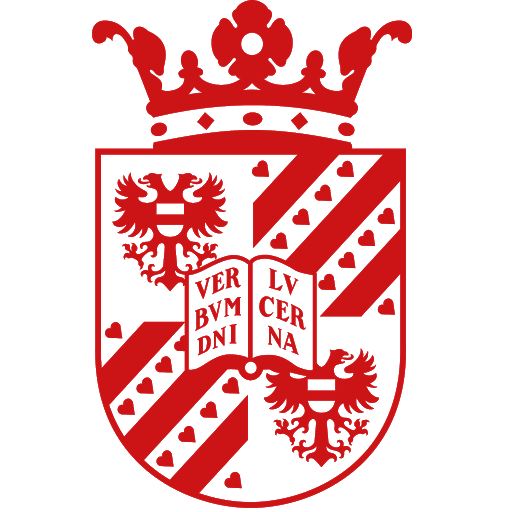}\\
  \vspace{1cm}
  {\large July 2021}\\[1cm]
\end{titlepage}

\setlength{\headheight}{0pt}
\tableofcontents
\addtocontents{toc}{~\hfill\textbf{Page}\par}

\newpage
\section*{Abstract}
\label{sec:abstract}
\addcontentsline{toc}{section}{Abstract}
The human brain is capable of learning, memorizing, and regenerating a panoply of temporal patterns. A neuro-dynamical mechanism called \textit{conceptors} offers a method for controlling the dynamics of a recurrent neural network by which a variety of temporal patterns can be learned and recalled. However, conceptors are matrices whose size scales quadratically with the number of neurons in the recurrent neural network, hence they quickly become impractical. In the work reported in this thesis, a variation of conceptors is introduced, called \textit{diagonal conceptors}, which are diagonal matrices, thus reducing the computational cost drastically. It will be shown that diagonal conceptors achieve the same accuracy as conceptors, but are slightly more unstable. This instability can be improved, but requires further research. Nevertheless, diagonal conceptors show to be a promising practical alternative to the standard full matrix conceptors.

\newpage
\section{Introduction}
\label{sec:introduction}
In 1997, the world chess champion Garry Kasparov lost to the IBM supercomputer Deep Blue. This was the first time a world chess champion was beaten by a machine \cite{AIBook1}. Ever since, the field of Artificial Intelligence (AI) has expanded and grown immensely. Currently, there are several books which are solely devoted to the discussion about the progress of AI, both the good and bad sides of it \cite{AIBook1,AIBook2}. One important open question is how close AI research is to creating AI that can outperform humans in most tasks. At present, AI can transcend human intelligence in some individual tasks, such as playing games like Atari, chess, and Go \cite{MuZero}, or arithmetic. Furthermore, the field of neural networks has made an impressive leap forward since its creation, instigated by the perceptron \cite{Perceptron}. At present, neural networks are researched extensively and are used for a wide range of tasks. Nevertheless, the human brain remains overall superior to AI. One of the reasons is that the human brain is able to learn a panoply of skills consisting of, for instance language, sport, and social interaction. Specifically, the ability to learn, recognize, recall, and combine a vast amount of \textit{temporal} patterns with little effort. It is thus desirable to research a neural network that is capable of such tasks. One of the difficulties is managing long-term memory, which is the ability to permanently store temporal patterns in its synaptic weights such that it is able to recall the learned patterns whenever asked without changing those weights, also referred to as neural long-term memory. There are several researches devoted to this task, which will be discussed in Section \ref{sec:related_work}. However, a neuro-computational mechanism called \textit{conceptors}, not only offers a solution to managing long-term memory for temporal patterns, but also offers insight into a variety of other problems \cite{MonsterReport}.
\doubleEnter
The conceptors architecture is a neuro-computational mechanism that can be used to control the dynamics of an \textit{Recurrent Neural Network} (RNN), henceforth also referred to as the \textit{reservoir}. Conceptors were first introduced in the technical report \textit{Controlling Recurrent Neural Network by Conceptors} written by H. Jaeager, which will be referred to as the \textit{conceptors report}. In the conceptors report many of the applications of conceptors, such as temporal pattern classification, human motion generation, de-noising and signal separation were demonstrated \cite{MonsterReport}. Consequentially, conceptors have been successfully used in several other researches \cite{ConceptorExample1, ConceptorExample2, ConceptorExample3, ConceptorExample4, ConceptorExample5, ConceptorExample6, RFCBirdsongs, RFCBistablePerception}. It must be noted that conceptors can take different forms, but in this report they will only be in matrix form and will therefore be referred to as \textit{conceptor matrices} or simply \textit{conceptors}.
\doubleEnter
Intuitively, conceptors act as filters through which an RNN can be controlled. The crucial observation is that when an $N$-dimensional RNN is driven by different patterns $1, 2,..., p$, different areas $A^{1}, A^{2},...,A^{p}$ of the neural state space are being activated. Conceptors exploit this observation, which allows for retrieving a desired pattern from the neural state space. Differently stated, conceptors attempt to identify the subvolumes in which the respective patterns live. A conceptor $C^{j}\in\mathbb{R}^{N\times N}$ representing a pattern $j$ constrains the neural dynamics to the volume of state space $A^{j}$ such that the network will regenerate pattern $j$. It achieves this by leaving the area of state space associated with pattern $j$ for the most part unchanged, yet suppressing the other areas of state space. Conceptors are robust against small parameter changes and noise. However, conceptors can become computationally inefficient. As the number of neurons in the reservoir $N$ increases, the conceptor matrices grow quadratically in $N$, which poses problems with storage and computational cost. For example, a matrix with $1000$ rows and columns uses approximately $8$ MB of storage, so for $10$ patterns, $80$ MB is needed. However, if the number of neurons in the reservoir increases from $N=1000$ to $N=5000$, then the conceptor matrix will have $5000$ rows and columns, requiring $200$ MB of memory for a single conceptor. Not only storage, but also the computational cost increases drastically. If $C\in\mathbb{R}^{N\times N}$ and $\mathbf{x}\in\mathbb{R}^{N}$, then the complexity of the multiplication $C\mathbf{x}$ is $\mathcal{O}(N^{2})$. Therefore, conceptor matrices quickly become unpractical as the reservoir size increases, which is unfortunate, as conceptors carry many useful qualities and it would be ideal if they could be used in practice. An alternative architecture was introduced, called the \textit{Random Feature Conceptor} (RFC) architecture, to overcome this impracticality as well as biological implausibility \cite{MonsterReport}. The idea of the RFC architecture is to expand the state of the RNN onto a higher-dimensional space, where the state is manipulated by a diagonal conceptor matrix rather than a full conceptor matrix, and finally the manipulated state is projected back to the original space. In this architecture, the diagonal conceptors manipulate the higher-dimensional state element-wise, so it can be written in vector form. This is computationally much more efficient. During the research conducted for this thesis it was discovered that the architecture still works if the dimension of the higher-dimensional space is set equal to dimension of the original space and both projection matrices are set to the identity matrix, i.e. the same architecture as conceptors, but with a diagonal conceptor matrix. However, this only works well if a vital adjustment is made in the training scheme, which will be discussed later. This discovery lead to the neuro-computational mechanism that is the subject of this thesis, called \textit{diagonal conceptors}.
\doubleEnter
Diagonal conceptors offer a solution to the impracticality of conceptors. As can be deduced from the name, diagonal conceptors are diagonal matrices, meaning that they reduce the computational cost significantly and require much less storage. A diagonal conceptor matrix $D\in\mathbb{R}^{N\times N}$ can be written as $D=\text{diag}(\mathbf{c})$, where $\text{diag}(.)$ denotes a diagonal matrix with zeros everywhere except for the diagonal. The diagonal $\mathbf{c}$ is an $N$-dimensional vector called the \textit{conception vector} and its elements $c_{i}$, for $i=1,...,N$, are called the \textit{conception weights}, where the terminology is borrowed from Section 3.14 of the conceptors report \cite{MonsterReport}. In practice, the conception vector $\mathbf{c}$ is used, but in this report, for notational simplicity, the matrix notation $D$ will be used. In comparison to a conceptor matrix, a conception vector of size $5000$ requires only $40$ KB of storage, so for $10$ patterns, only $0.4$ MB is required to store the conception vectors. Moreover, the complexity of computing the conception vector $\mathbf{c}$ times a vector $\mathbf{x}\in\mathbb{R}^{N}$ is only $\mathcal{O}(N)$, which is a considerable decrease compared to $\mathcal{O}(N^{2})$.
\doubleEnter
In addition, it will be shown that the conception weights can be trained individually, which provides two advantages over conceptors. First, diagonal conceptors are not biologically implausible, whereas conceptors are. Conceptors can be online adapted in a variation called \textit{autoconceptors}, but the online adaptation requires non-local computations, making the conceptors biologically implausible. Diagonal conceptors, on the other hand, can also be adapted online, for which only local computations are required. Therefore, it can be said that diagonal conceptors are not biologically implausible, in the sense that all information for the adaptation of a synaptic weight is available as the synapse. Second, the global learning rate for conceptors is constrained by the ratio of the smallest to the largest curvature direction in the gradient space, resulting in slow convergence in areas where the curvature is small. The learning rate of the online adaptation of diagonal conceptors, on the other hand, can be chosen for each conception weight individually. Therefore, the constraint of the global learning rate for conceptors is lifted for diagonal conceptors.
\doubleEnter
Diagonal conceptors are trained differently than conceptors, which is not surprising, since the degrees of freedom of a conceptor is much higher than the degrees of freedom of a diagonal conceptor. The main change in the training scheme is that before the reservoir is driven, a randomly initialized diagonal conceptor is inserted in the update loop. These initial random diagonal conceptors randomly scale each neuron in the reservoir individually. Consequently, the different areas $A^{1}, A^{2},...,A^{p}$ of the neural state space that are being activated by driving the RNN with patterns $1,2,...,p$ are randomly scaled, which creates new areas $\hat{A}^{1}, \hat{A}^{2},...,\hat{A}^{p}$. The diagonal conceptors will be trained on those newly positioned areas $\hat{A}^{j}$ rather than the original areas $\hat{A}^{j}$, for all $j$. This is especially useful when different areas in state space overlap, since, in contrast to conceptors, diagonal conceptors have less degrees of freedom than conceptors to characterize the nuances of each area. The use of this initial random scaling is an interesting observation and it offers opportunities for more future work, as it could be researched whether the randomness of the initial random scaling can be optimized.
\doubleEnter
This report introduces the diagonal conceptors by examples that are also used in the conceptors report \cite{MonsterReport}. The examples consist of \textit{four periodic patterns}, \textit{chaotic attractors}, and \textit{human motions}. Since each example highlights a different quality of conceptors, it will become evident which characteristics translate to diagonal conceptors and which do not. For each example, both conceptors and diagonal conceptors are trained, allowing a direct comparison to be made.
\doubleEnter
The objective of this report is to introduce diagonal conceptors so they can be used as a practical alternative for matrix conceptors. Furthermore, this report aims to analyse some of the properties of diagonal conceptors and how they relate the properties of conceptors. Lastly, this report is intended as a intuitive guide to using diagonal conceptors in practice.

\newpage
\section{Related Work}
\label{sec:related_work}
The diagonal conceptors architecture was inspired by the RFC architecture, which is described in Section $3.15$ of the conceptors report \cite{MonsterReport}. Therefore, diagonal conceptors are closely related to RFCs and many of the consequences of RFCs translate to diagonal conceptors. However, it should be noted that RFCs are trained differently than diagonal conceptors. The conception weights in the RFC architecture are trained using an adaptation rule that is evaluated every time step during an adaption period, whereas the conception weights in the diagonal conceptors architecture are computed explicitly after a state collection period. Consequently, the conception weights in the RFC architecture are restricted to either $0$ or the range $(0.5,1]$, whereas the conception weights in the diagonal conceptors architecture are restricted to the broader range $(0,1)$. This difference in the resulting conception weights changes the dynamics of the reservoir, but it does not affect the translation of the properties of RFCs to diagonal conceptors. For example, diagonal conceptors are also computationally efficient and not biologically implausible, like RFCs, which can be argued by the same reasoning as in Section 3.15 of the conceptors report \cite{MonsterReport}. Furthermore, the algebraic and logical rules for conception weights also translate to diagonal conceptors. The symbolic interpretation of conceptors is not discussed in this thesis, so an analysis on the algebraic and logical rules for the conception weights of diagonal conceptors is an opportunity for future work.
\doubleEnter
RFCs have remained mostly unexplored in literature. As of yet, there are only two articles that make use of RFCs. The first article uses RFCs in their model to learn and recognize complex, dynamic stimuli. They conclude that RFCs perform well, but are prone to instability \cite{RFCBirdsongs}. The second used the RFC architecture to accurately model bistable perception. They conclude that the RFC architecture is a promising model for general human perception \cite{RFCBistablePerception}. The almost nonexistence of RFCs in the literature may be attributed to the unexplored shortcomings of the model. As was pointed out in one of the articles, it can be difficult for RFCs to converge to a stable solution. In the conceptors report, this was pointed out as well \cite{MonsterReport}. In the examples in this report, diagonal conceptors are shown to yield stable reservoir dynamics, which offers a promising approach to applying conceptors in practice compared to RFCs.
\doubleEnter
Other related research can be divided up in two parts. The first is by viewing diagonal conceptors as a variation of conceptors. The scope of the conceptors architecture is much broader than learning, recognizing, and recalling temporal patterns. Conceptors were introduced as a novel perspective on the neuro-symbolic integration problem\cite{MonsterReport}. The field of neuro-symbolic integration aims to bridge the gap between two fundamentally different paradigms: symbolic systems in AI and neural network systems in AI. In turn, it intents to unite the research of different fields. There are many researches from different fields dedicated to the neuro-symbolic integration problem \cite{NeuroSymbolic1, NeuroSymbolic3, NeuroSymbolic4, NeuroSymbolic6}. However, the field of neuro-symbolic integration is outside the scope of this report, so for an easy introduction the interested reader is referred to \cite{NeuroSymbolicIntegrationBook}. As diagonal conceptors are only used for storing and recalling temporal patterns in this report, the neuro-integration problem will not be discussed.
\doubleEnter
The second way of comparing diagonal conceptors to other works is by viewing the diagonal conceptors architecture as a means to store temporal patterns in neural long-term memory. The article \textit{Using Conceptors to Manage Neural Long-Term Memories for Temporal Patterns} by H. Jaeger \cite{ConceptorsHumanMotion}, shows how conceptors can be used to manage neural long-term memories for temporal patterns. In its introduction, a variety of different models and techniques is discussed that have attempted the same. First, associative memory is considered, which is the paradigmatic model for neural long-term memory. It was introduced by a variety of researches \cite{AutoAssociative1,AutoAssociative2,AutoAssociative3}, only a few of which are cited here. However, these models are mostly designed for \textit{static} patterns, such as images, and not for temporal patterns. Second, it considers a number of approaches for managing neural memories of temporal patterns, a few of which are an extension of the classical associative memory model. It consists of, but is not limited to: hetero-associative memory \cite{RelatedWorkHetero}, multiassociative memory \cite{RelatedWorkMultiassociative}, reservoir computing \cite{RelatedWorkReservoirComputing}. These approaches will not be discussed in detail here, as they are discussed in detail in \cite{ConceptorsHumanMotion}. However, the reason they are briefly mentioned here is that they are compared to conceptors in a later part of the introduction of \cite{ConceptorsHumanMotion}. In this part, a number of properties that are variously attributed to a neural memory system are listed. Each of the mentioned approaches can be attributed a number of properties, but usually they fall short in the others. Conceptors, on the other hand, facilitate all those properties except two. Now, since the argument in favor of conceptors can be translated to diagonal conceptors it follows that diagonal conceptors also facilitate all those properties except the two. Moreover, compared to conceptors, diagonal conceptors have the advantage of computational efficiency as well as the fact that diagonal conceptors are not biologically implausibility. Therefore, for all the advantages of conceptors over the discussed other methods in \cite{ConceptorsHumanMotion}, diagonal conceptors possess the same advantages plus two more.

\newpage
\section{Theory} 
\label{sec:theory}
The first part of this section largely follows the conceptors report \cite{MonsterReport}. Nevertheless, this section is self-contained and introduces all the required background knowledge for this thesis. It is organized in four parts. First and second, the mathematical landscape of an RNN is introduced as well as how to store patterns in a reservoir and how to train the output weights. Third, conceptors are introduced and defined, followed by a brief introduction of autoconceptors. Fourth, diagonal conceptors are introduced and defined.
\doubleEnter
As for mathematical notation, there are a few notations that will be consistent throughout this report. A matrix is denoted by a capital letter. A vector is denoted by a bold letter. The element on the $i$-th row and $j$-th column of a matrix $A$ is denoted by $a_{ij}$. The $i$-th element of a vector $\mathbf{a}$ is denoted by $a_{i}$. The transpose of a matrix $A$ is denoted by $A^{T}$. The Frobenius norm of a matrix $A$ is denoted by $||A||_{fro}$ and is defined, for a real matrix, as $||A||_{fro}=\sqrt{\sum_{i}\sum_{j}|a_{ij}|^{2}}$. If the norm has no subscript, i.e. $||A||$, the reader can assume that the Frobenius norm is meant.

\subsection{Mathematical Landscape} \label{sec:mathematical_landscape}
\begin{figure}[t]
    \centering
    \includegraphics[width=0.7\textwidth]{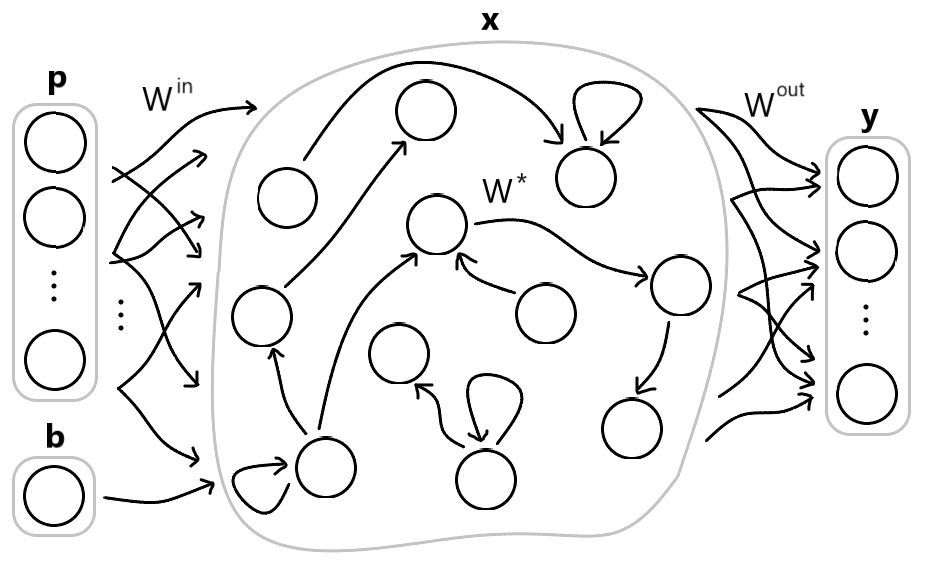}
    \caption{Basic setup of the RNN. The input neurons drive the reservoir with pattern $\mathbf{p}$ through an input weights matrix $W^{in}$ as well as a bias vector $\mathbf{b}$. The reservoir neurons are interconnected via a reservoir weights matrix $W^{*}$. The output $\mathbf{y}$ can be read from the reservoir state $\mathbf{x}$ via an output weights matrix $W^{out}$.}
    \label{fig:RNN}
\end{figure}

As mentioned in the introduction, the goal is to store a variety of temporal patterns in a single RNN such that they can later be retrieved. With \textit{storing} it is meant that after the patterns are stored in the RNN, the RNN can autonomously regenerate any pattern. Furthermore, \textit{retrieving} a pattern from an RNN means that the autonomous regeneration of any stored pattern can be instigated at any time such that the desired pattern is outputted from the RNN. The RNN will now be made explicit.
\doubleEnter
Let $\mathbf{P}=\{\mathbf{p}^{1},\mathbf{p}^{2},...,\mathbf{p}^{p}\}$ be a set of $p$ discrete-time patterns, where pattern $j$ at time step $n$ is given by $\mathbf{p}^{j}(n)\in\mathbb{R}^{M}$. Furthermore, assume $M$ input neurons as well as a bias neuron, an RNN consisting of $N$ simple tanh neurons called \textit{reservoir neurons}, and $M'$ output neurons. In the more general case one can assume $M'\neq M$ output neurons, which would be the case if the driving patterns are used to produce output based on the driving patterns, e.g. classification. However, in this report the reservoir and output neurons serve as a means to self-generate the driving patterns, hence the number of input and output neurons is set equal. The input neurons drive the reservoir with $\mathbf{p}^{j}(n)$ via an input weights matrix $W^{in}\in\mathbb{R}^{N\times M}$ and a bias vector $\mathbf{b}\in\mathbb{R}^{N}$. The $N$ neurons in the RNN are recurrently connected via a weights matrix $W^{*}\in\mathbb{R}^{N\times N}$. When the reservoir is driven by pattern $\mathbf{p}^{j}(n)$, the reservoir neurons are activated. The state of neuron $i$ at time $n$ is denoted by $x_{i}^{j}(n)\in(-1,1)$ and is called the \textit{neuron state}. The state of the reservoir at time $n$, denoted by $\mathbf{x}^{j}(n)\in(-1,1)^{N}$, is given by the vector containing all the neuron states at time $n$ and is called the \textit{reservoir state}. Finally, the output neurons serve to read out the target signal at time step $n$, denoted by $\mathbf{y}^{j}(n)\in\mathbb{R}^{M}$, from the reservoir state via the output weights matrix $W^{out}\in\mathbb{R}^{M\times N}$. A visual representation of the setup can be seen in Figure \ref{fig:RNN}.
\doubleEnter
Let $W^{*}$, $W^{in}$, and $\mathbf{b}$ be fixed random matrices. They will not be adjusted after initialization. The reservoir state vector, or simply \textit{state vector}, is updated according to the update equation
\begin{equation}
\label{eq:mathematical_landscape_update_equation}
    \mathbf{x}^{j}(n+1)=\text{tanh}(W^{*}\mathbf{x}^{j}(n)+W^{in}\mathbf{p}^{j}(n+1)+\mathbf{b})
\end{equation}
and the output signal is given by
\begin{equation}
\label{eq:mathematical_landscape_output_equation}
    \mathbf{y}^{j}(n)=W^{out}\mathbf{x}^{j}(n),
\end{equation}
where the output weights $W^{out}$ are learned, see Section \ref{sec:theory_storing_patterns}. Let the reservoir be driven for $L$ time steps. The state vectors $\mathbf{x}^{j}(n)$ are collected in the \textit{state collection matrix} $X^{j}$. The state collection matrix gives a sample of states that is representative of the volume of state space that is occupied by the driving pattern $j$ and is given by
\begin{equation}
    X^{j}=
    \begin{bmatrix}
    x_{1}^{j}(1) & x_{1}^{j}(2) & \hdots & x_{1}^{j}(L) \\
    x_{2}^{j}(1) & x_{2}^{j}(2) & \hdots & x_{2}^{j}(L) \\
    \vdots &  & \ddots & \vdots \\
    x_{N}^{j}(1) & x_{N}^{j}(2) & \hdots & x_{N}^{j}(L)
    \end{bmatrix}.
\end{equation}
Note that the information about pattern $j$ is encoded in the $X^{j}$.

\subsection{Storing the Patterns and Training the Output Weights}
\label{sec:theory_storing_patterns}
Up until this point, the reservoir has been driven by the input patterns. However, in order to store the patterns, the reservoir must be able to update the state vector in the absence of a driving input, which yields the following approximation:
\begin{equation}
    \text{tanh}(W^{*}\mathbf{x}^{j}(n)+W^{in}\mathbf{p}^{j}(n+1)+\mathbf{b})\approx\text{tanh}(W\mathbf{x}^{j}(n)+\mathbf{b}),
\end{equation}
where $W$ comprises the \textit{recomputed reservoir weights}. $W$ can be computed by minimizing the mean square error over all $j$ and $n$, yielding
\begin{equation}
W = \underset{\Tilde{W}}{\mathrm{argmin}}\sum_{j}\sum_{n}||W^{*}\mathbf{x}^{j}(n)+W^{in}\mathbf{p}^{j}(n+1)-\Tilde{W}\mathbf{x}^{j}(n)||^{2}.    
\end{equation}
For such linear regression tasks, ridge regression will be used throughout this report. Furthermore, it should be noted that there are alternative techniques for removing the driving input from the update equations, which are discussed in Section 3.11.1 of the conceptors report \cite{MonsterReport}.
\doubleEnter
For computing $W^{out}$, a similar approach is employed, where the output $\mathbf{y}^{j}(n)=W^{out}\mathbf{x}^{j}(n)$ must approximate the input pattern $\mathbf{p}^{j}(n)$. This gives the approximation
\begin{equation}
    \mathbf{p}^{j}(n)\approx\mathbf{y}^{j}(n)=W^{out}\mathbf{x}^{j}(n),
\end{equation}
where $W^{out}$ is then computed by minimizing the mean square error over all $j$ and $n$, which yields
\begin{equation}
    W^{out} = \underset{\Tilde{W}^{out}}{\mathrm{argmin}}\sum_{j}\sum_{n}||\mathbf{p}^{j}(n)-\Tilde{W}^{out}\mathbf{x}^{j}(n)||^{2}.
\end{equation}
The process of recomputing the reservoir weights $W$ and computing the output weights $W^{out}$ is referred to as \textit{storing} the patterns in the reservoir. The reservoir is called \textit{loaded} after the patterns have been stored.

\subsection{Conceptors}
\label{sec:theory_conceptors}
\begin{figure}[t]
    \centering
    \includegraphics[width=0.8\textwidth]{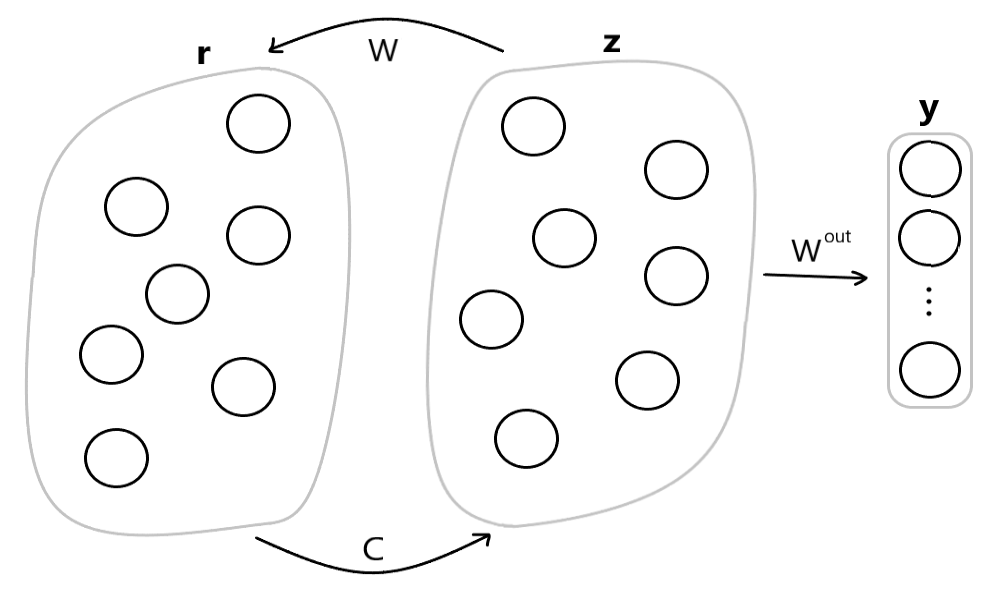}
    \caption{A visual representation of Equations \ref{eq:theory_conceptors_update_equation} and \ref{eq:theory_conceptors_output}. The state $\mathbf{r}$ is projected to a state $\mathbf{z}$ via a conceptor matrix $C$ and $\mathbf{z}$ is inserted back in the loop via $W$. The output is read from the state $\mathbf{z}$ by the output weights $W^{out}$.}
    \label{fig:RNN2}
\end{figure}

After the patterns are stored in the reservoir, there exists a superposition of patterns in the reservoir. If the network were to simply start updating the state vector by $\mathbf{x}(n+1)=\tanh{(W\mathbf{x}(n)+\mathbf{b})}$ it would exhibit unpredictable behavior, since the reservoir does not know which pattern to engage in. So, how is a specific pattern retrieved from the reservoir? Note that driving the reservoir with pattern $j$ creates a cloud of points in state space, which is characteristic for pattern $j$. This point cloud is given by the columns of the state collection matrix $X^{j}$. Ideally, the point cloud associated with pattern $j$ is confined to a proper subspace $S^{j}\subset\mathbb{R}^{N}$ such that the state vector can be projected on $\mathcal{S}^{j}$ via a projection matrix $\mathbf{P}_{S^{j}}$. However, the state vectors usually span the entire space $\mathbb{R}^{N}$, so $\mathbf{P}_{\mathbb{R}^{N}}$ would be the identity matrix, which would not be helpful in singling out an individual pattern. Therefore, instead of applying the projection $\mathbf{P}_{\mathbb{R}^{N}}$, the state vector is projected on a subspace that is spanned by a number of leading principal components of the point cloud associated with pattern $j$. How many of the leading principal components should be used in this soft projection is adjustable by means of a control parameter called \textit{aperture}\footnote{The name aperture has its roots in optics, where it means the diameter of the effective lens opening, hence it negotiates the amount of light energy that reaches the film.}. Each pattern $j$ will have its own projection matrix, which is called the \textit{conceptor} associated with pattern $j$ and is denoted by $C^{j}$. The conceptor matrix $C^{j}$ should behave such that it leaves states associated with pattern $j$ intact, but suppresses the states of other patterns. This trade-off is mitigated by the aperture.
\doubleEnter
Applying $C^{j}$ to the state vector can be viewed as a neural network consisting of two layers, where the first layer is projected to the second layer via $C^{j}$ and is projected back via $W$. Therefore, the update equation is written in two parts
\begin{equation}
\label{eq:theory_conceptors_update_equation}
\begin{aligned}
    \mathbf{r}(n+1)&=\tanh{(W\mathbf{z}(n)+\mathbf{b})},\\
    \mathbf{z}(n+1)&=C^{j}\mathbf{r}(n+1),
\end{aligned}
\end{equation}
where the output is given by
\begin{equation}
\label{eq:theory_conceptors_output}
    \mathbf{y}(n)=W^{out}\mathbf{z}(n).
\end{equation}
A visual representation is depicted in Figure \ref{fig:RNN2}.

\subsubsection{Computing Conceptors}
Conceptor $C^{j}$ must leave states $\mathbf{z}^{j}(n)$ unchanged, but also suppress components that are not typical for $\mathbf{z}^{j}(n)$. This essentially means that $C^{j}$ should act as the identity matrix for states $\mathbf{z}^{j}(n)$, but as the null matrix for the unwanted components. This leads to the following quadratic loss function:
\begin{equation}
\begin{aligned}
    \mathcal{L}(C^{j})&=\mathbb{E}[||\mathbf{z}^{j}-C^{j}\mathbf{z}^{j}||^{2}] + \alpha^{-2}||C^{j}||^{2}\\
    &=\text{Tr}[(I-(C^{j})^{T})(I-C^{j})]\mathbb{E}[\mathbf{z}^{j}(\mathbf{z}^{j})^{T}]+(\alpha^{j})^{-2}\text{Tr}[(C^{j})^{T}C^{j}],
\end{aligned}
\end{equation}
where $\alpha^{j}$ is the aperture associated with pattern $j$ and $||.||$ is the Frobenius norm.
\doubleEnter
The first component of $\mathcal{L}(C^{j})$ is the time-average of the difference between the projected state vectors $C^{j}\mathbf{z}^{j}$ and the state vector $\mathbf{z}^{j}$. This reflects that $C^{j}$ must leave $\mathbf{z}^{j}$ unchanged. This component is optimal for $C^{j}=I$. The second part of the loss function represents the suppression of the outlying components of the state vector. Note that it is minimal for $C^{j}$ equals the null matrix. Therefore, a large aperture results in a conceptor matrix close to the identity matrix. Conversely, a small aperture will shrink the conceptor matrix towards the null matrix.
\doubleEnter
To find the matrix $C^{j}$ for which $\mathcal{L}(C^{j})$ is minimal, the gradient of $\mathcal{L}(C^{j})$ with respect to $C^{j}$ is computed. The gradient is given by
\begin{equation}
\label{eq:L_derivative}
    \frac{\partial}{\partial C^{j}}\Big(\mathbb{E}[||\mathbf{z}^{j}-C^{j}\mathbf{z}^{j}||^{2}] + (\alpha^{j})^{-2}||C^{j}||^{2}\Big)=-2(I-C^{j})\mathbb{E}[\mathbf{z}^{j}(\mathbf{z}^{j})^{T}]+2(\alpha^{j})^{-2}C^{j}.
\end{equation}
It is then straightforward to see that the minimization of $\mathcal{L}(C^{j})$ leads to the following solution
\begin{equation}
\label{eq:compute_Cj}
    C^{j}=R^{j}(R^{j}+(\alpha^{j})^{-2}I)^{-1},
\end{equation}
where $R^{j}=\mathbb{E}[\mathbf{z}^{j}(\mathbf{z}^{j})^{T}]$ is the \textit{state correlation matrix} associated with pattern $j$. For $\alpha^{j}\in(0,\infty)$, the solution is well-defined, but for $\alpha^{j}=0$ it is not. The cases of $\alpha^{j}=0$ and $\alpha^{j}=\infty$ a will not be discussed here, but they are discussed in Section 3.8.1 of the conceptors report \cite{MonsterReport}. In simulations, the state correlation matrix $R^{j}$ is estimated by $\hat{R}^{j}=Z^{j}(Z^{j})^{T}/L$, where $Z^{j}=[\mathbf{z}^{j}(1)\;\mathbf{z}^{j}(2)\;...\;\mathbf{z}^{j}(L)]\in\mathbb{R}^{N\times L}$ is the state collection matrix from Section \ref{sec:mathematical_landscape}.
\doubleEnter
A few properties of $C^{j}$ and $R^{j}$ can be inferred and are worth mentioning:
\begin{enumerate}
    \item $C^{j}$ and $R^{j}$ have the same $N$ eigenvectors, meaning that if $R^{j}=U\Sigma U^{T}$ is the singular value decomposition of $R^{j}$ then $C^{j}=USU^{T}$, where $\Sigma$ and $S$ contain the singular values of $R^{j}$ and $C^{j}$, respectively.
    \item The singular values of $C^{j}$, denoted by $s_{i}^{j}$, are the normalized singular values of $R^{j}$, denoted by $\sigma_{i}^{j}$, and relate by $s_{i}^{j}=\sigma_{i}^{j}(\sigma_{i}^{j}+(\alpha^{j})^{-2})^{-1}$.
    \item From the previous property it can be concluded that $0\leq s_{i}^{j}\leq 1$.
\end{enumerate}

\subsubsection{Morphing}
\label{sec:conceptors_morphing}
The dynamics of the reservoir can be morphed by conceptors. The term morphing is usually used to describe the smooth transition from one image to another by slow gradual interpolation, but here is it used to describe a smooth transition between reservoir dynamics. Conceptors can be used to morph different patterns that are stored in the reservoir. Let the reservoir be loaded with patterns $j_{1}$ and $j_{2}$ and let the corresponding conceptors $C^{j_{1}}$ and $C^{j_{2}}$ be given. Then, a mixture of the reservoir dynamics associated with $\mathbf{p}^{j_{1}}$ and $\mathbf{p}^{j_{2}}$ can be obtained by a linear combination of $C^{j_{1}}$ and $C^{j_{2}}$, given by
\begin{equation}
\label{eq:conceptors_morphing}
\begin{aligned}
    \mathbf{r}(n+1)&=\tanh{(W\mathbf{z}(n)+\mathbf{b})},\\
    \mathbf{z}(n+1)&=\big((1-\mu)C^{j_{1}} + \mu C^{j_{2}}\big)\mathbf{r}(n+1),
\end{aligned}
\end{equation}
where $\mu\in\mathbb{R}$ is the \textit{mixture parameter}. Notice that the mixture parameter $\mu$ is not constrained to $[0,1]$. As will be shown in the simulations in Section \ref{sec:simulations_morphing}, conceptors are not only able to interpolate, but also extrapolate. In the conceptors report, it was shown that for two sine waves with different periods that conceptors are capable of capturing the period of the sine waves and then not only interpolate between them, but also extrapolate \cite{MonsterReport}. The periods in the example were $\approx8.83$ and $9.83$, but with the extrapolation, the conceptors were able to create sine waves of periods between $\approx7.5$ and $\approx11.9$.
\doubleEnter
Furthermore, it must be noted that a morph is not restricted to two patterns. In the more general case, a morph can be performed for $p$ patterns, where the term $(1-\mu)C^{j_{1}} + \mu C^{j_{2}}$ would be substituted by $\sum_{i=1}^{p}\mu_{i}C^{j_{i}}$. In this sum, the mixing parameter $\mu_{i}$ determines how much the outputted pattern is influenced by the dynamics of pattern $j_{i}$. The simulations in this report only make use of the case where $p=2$.

\subsubsection{Autoconceptors}
\label{sec:theory_conceptors_autoconceptors}
Thus far, conceptors are computed with Equation \ref{eq:compute_Cj}, after which they are stored so the dynamics of the reservoir can be constrained at a later time. This is practical and achievable for machine learning applications, where the conceptors can be written to a file. However, this is not always the case. Specifically, from a neuroscience point of view, it seems unlikely that a filter is created for every new pattern that needs to be learned, where the filter has the same size as the reservoir. This compels for a different architecture, where the conceptors are created while the reservoir is being driven and the conceptors need not be stored.
\doubleEnter
The idea is to make a conceptor matrix time-dependent, so it can be adapted while the reservoir is being driven. The update equations in Equation \ref{eq:theory_conceptors_update_equation} then change to
\begin{equation}
\label{eq:theory_autoconceptors_update_equation}
\begin{aligned}
    \mathbf{r}(n+1)&=\tanh{(W\mathbf{z}(n)+\mathbf{b})},\\
    \mathbf{z}(n+1)&=C(n)\mathbf{r}(n+1),
\end{aligned}
\end{equation}
where the only difference is that $C$ is now time-dependent. The adaptation rule for the online adaptation of $C(n)$ can be read directly from Equation \ref{eq:L_derivative}, yielding
\begin{equation}
    C(n+1) = C(n) + \lambda\Big( \big(I-C(n)\big)\mathbf{z}(n)\mathbf{z}^{T}(n) - \alpha^{-2}C(n)\Big),
\end{equation}
where $\lambda>0$ is the learning rate. Note that $\lambda$ is constrained by the ratio of the smallest to the largest curvature direction in the gradient space. This leads to slow convergence in areas where the local curvature is lower.
\doubleEnter
It was shown that if the autoconceptor $C(n)$ converges, it will possess the same algebraic properties as a conceptor \cite{MonsterReport}. Autoconceptors comprise a large part of the conceptors report, so for more details the reader is referred there.

\subsection{Diagonal Conceptors}
\label{sec:theory_diagonal_conceptors}
\begin{figure}[t]
    \centering
    \includegraphics[width=0.8\textwidth]{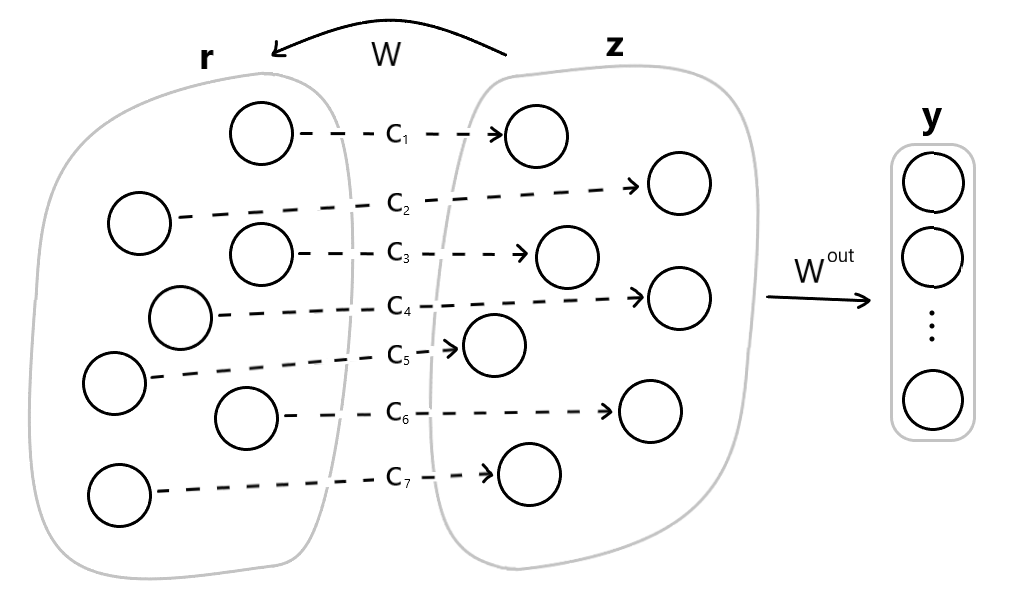}
    \caption{A visual representation of Equations \ref{eq:diagonal_conceptors_output_equation} and \ref{eq:diagonal_conceptors_update_equation_vector}. The state $\mathbf{r}$ is element-wise scaled to the state $\mathbf{z}$. This is denoted by the dashed lines, where each dashed line depicts a scaling by a conception weight $c_{i}$. The state $\mathbf{z}$ is inserted back in the loop via $W$. The output is read from the state $\mathbf{z}$ by the output weights $W^{out}$.}
    \label{fig:RNN3}
\end{figure}

There are two main areas in which conceptors fall short. First, conceptors are computationally expensive. As mentioned before, each pattern that needs to be stored in the reservoir, also requires a conceptor matrix, which has the same size as the reservoir. Furthermore, the conceptor matrices quickly increase in size as they scale quadratically with the number of reservoir neurons $N$. Therefore, conceptors could hardly be used in real-world applications where the dimension of the reservoir is large. Second, conceptors are biologically implausible. The online adaptation of an element of the conceptor matrix, denoted by $C_{ij}$, requires information that would biologically not all be available at the synapse of $C_{ij}$.
\doubleEnter
In Section 3.15 of the conceptors, an architecture, called \textit{Random Feature Conceptors} (RFC), is introduced that solves the shortcomings of conceptors \cite{MonsterReport}. In short, the idea of random feature conceptors is to project the reservoir state to a higher-dimensional space, where it is manipulated by a diagonal conceptor matrix rather than a full conceptor matrix, and finally the manipulated state is projected back to the original space. In this architecture, the diagonal matrix conceptors manipulate the higher-dimensional state element-wise, so it can be written in vector form. This is computationally much more efficient and is not biologically implausible, which is explained in more detail in Section 3.15 of the conceptors report \cite{MonsterReport}. 
\doubleEnter
Here, an architecture, called \textit{diagonal conceptors}, is proposed that is closely related to RFC. In contrast to RFC, in the diagonal conceptors architecture, the projection to the higher-dimensional space is unnecessary. Therefore, the conceptor matrix $C^{j}$ in Equation \ref{eq:theory_conceptors_update_equation} can be simply substituted with a \textit{diagonal conceptor matrix} $D^{j}$. It will be shown that this architecture only works well if an adjustment is made in the training algorithm of conceptors, which will be made clear in Section \ref{sec:training_diagonal_conceptors}. This section discusses how the diagonal conceptors would theoretically be computed.
\doubleEnter
A diagonal matrix is a matrix in which all the off-diagonal entries are zero. Let $\mathbf{c}=[c_{1}\;c_{2}\;...\;c_{N}]\in\mathbb{R}^{N}$, then a diagonal matrix $D\in\mathbb{R}^{N\times N}$ is often denoted by
\begin{equation}
    D=\text{diag}(\mathbf{c})=
    \begin{bmatrix}
    c_{1} & 0 & \hdots & 0\\
    0 & c_{2} & \hdots & 0\\
    \vdots & 0 & \ddots & \vdots\\
    0 & 0 & \hdots & c_{N}\\
    \end{bmatrix}.
\end{equation}
Let $D^{j}=\text{diag}(\mathbf{c}^{j})\in\mathbb{R}^{N\times N}$ denote the \textit{diagonal conceptor matrix}, or simply \textit{diagonal conceptor}, associated with pattern $j$, where $\mathbf{c}^{j}=[c_{1}^{j}\;c_{2}^{j}\;...\;c_{N}^{j}]\in\mathbb{R}^{N}$ is called the \textit{conception vector} and its elements $c_{i}^{j}$, for $i=1,2,...,N$, are called the \textit{conception weights}. Conception vector and conception weights are terminology that is borrowed from the RFC architecture, described in Section 3.15 of the conceptors report \cite{MonsterReport}. In Equation \ref{eq:theory_conceptors_update_equation}, the conceptor matrix $C^{j}$ is substituted for the diagonal conceptor matrix $D^{j}$, which yields
\begin{equation}
\label{eq:diagonal_conceptors_update_equation}
\begin{aligned}
    \mathbf{r}(n+1)&=\tanh{(W\mathbf{z}(n)+\mathbf{b})}\\
    \mathbf{z}(n+1)&=D^{j}\mathbf{r}(n+1)
\end{aligned}
\end{equation}
and the output can be read from the reservoir according to 
\begin{equation}
\label{eq:diagonal_conceptors_output_equation}
    \mathbf{y}(n)=W^{out}\mathbf{z}(n).
\end{equation}
Note that $\mathbf{z}(n)$ is now a scaling of $\mathbf{r}(n)$, where the scaling parameters are given by $\mathbf{c}^{j}$. A convenient way to write Equation \ref{eq:diagonal_conceptors_update_equation} is by means of the conception weights. This gives
\begin{equation}
\label{eq:diagonal_conceptors_update_equation_vector}
\begin{aligned}
    \mathbf{r}(n+1)&=\tanh{(W\mathbf{z}(n)+\mathbf{b})}\\
    \mathbf{z}(n+1)&=\mathbf{c}^{j}\circ\mathbf{r}(n+1),
\end{aligned}
\end{equation}
where $\circ$ is the Hadamard, or element-wise, multiplication operator. A visualization of Equations \ref{eq:diagonal_conceptors_output_equation} and \ref{eq:diagonal_conceptors_update_equation_vector} is shown in Figure \ref{fig:RNN3}, where a dashed line depicts an simple scaling operation. Simulations make use of Equation \ref{eq:diagonal_conceptors_update_equation_vector} rather than Equation \ref{eq:diagonal_conceptors_update_equation}, because $\mathbf{z}(n)$ can then be computed by vector-vector multiplication rather than matrix-vector multiplication.
\doubleEnter
Diagonal conceptors are computationally cheaper than conceptors, as they allow for vector storage and vector-vector multiplication. Furthermore, diagonal conceptors are not biologically implausible, which will be discussed further in Section \ref{sec:diagonal_autoconceptors}. However, it must be noted that the diagonal conceptors architecture does not imply biological plausibility. It merely suggest that it is not biologically implausible.

\subsubsection{Computing Diagonal Conceptors}
The derivation of a conceptor matrix can be translated directly to diagonal conceptors. Moreover, instead of deriving an expression for the diagonal conceptor $D^{j}$, an expression for the individual conception weights $c_{i}^{j}$ is derived. Similar to conceptor matrices, the diagonal conceptor associated with pattern $j$ must leave the states associated with pattern $j$ untouched, while suppressing the states associated with other patterns. Again, this balance is mitigated by the parameter \textit{aperture}, denoted by $\alpha^{j}$. This leads to a loss function similar to the loss function for a conceptor $C^{j}$. The loss function for diagonal conceptor $D^{j}$ is given by
\begin{equation}
\label{eq:theory_diagonal_conceptors_loss_function}
    \mathcal{L}(D^{j})=\mathbb{E}[||\mathbf{z}^{j}-D^{j}\mathbf{z}^{j}||^{2}] + (\alpha^{j})^{-2}||D^{j}||^{2},
\end{equation}
Since $D^{j}$ is a diagonal matrix, Equation \ref{eq:theory_diagonal_conceptors_loss_function} can be more conveniently written element-wise, which gives
\begin{equation}
\begin{aligned}
\label{eq:theory_diagonal_conceptors_loss_function_element_wise}
    \mathcal{L}(c_{i}^{j})&=\mathbb{E}[(z_{i}^{j}-c_{i}^{j}z_{i}^{j})^{2}] + (\alpha^{j})^{-2}(c_{i}^{j})^{2}\\
    &=(1-c_{i}^{j})^{2}\mathbb{E}[(z_{i}^{j})^{2}]+(\alpha^{j})^{-2}(c_{i}^{j})^{2}.
\end{aligned}
\end{equation}
An expression for $c_{i}^{j}$ is found by minimizing the loss function $\mathcal{L}(c_{i}^{j})$, for which the derivative with respect to $c_{i}^{j}$ must be computed. This gives
\begin{equation}
\label{eq:L_ci_derivative}
    \frac{\partial}{\partial c_{i}^{j}}\Big((1-c_{i}^{j})^{2}\mathbb{E}[(z_{i}^{j})^{2}]+(\alpha^{j})^{-2}(c_{i}^{j})^{2}\Big)=-2(1-c_{i}^{j})\mathbb{E}[(z_{i}^{j})^{2}]+2(\alpha^{j})^{-2}c_{i}^{j}.
\end{equation}
Setting this equal to $0$ and solving for $c_{i}^{j}$ immediately yields
\begin{equation}
\label{eq:theory_diagonal_conceptors_ci}
    c_{i}^{j}=\frac{\mathbb{E}[(z_{i}^{j})^{2}]}{\mathbb{E}[(z_{i}^{j})^{2}]+(\alpha^{j})^{-2}}.
\end{equation}
This expression indicates that $0<c_{i}^{j}<1$, where the boundaries $0$ and $1$ require a bit more attention. 
\doubleEnter
First, consider the value $\mathbb{E}[(z_{i}^{j})^{2}]$. Note that $\mathbb{E}[(z_{i}^{j})^{2}]$ can be written as $(c_{i}^{j})^{2}\mathbb{E}[(r_{i}^{j})^{2}]$, where $0<\mathbb{E}[(r_{i}^{j})^{2}]<1$, because of the hyperbolic tangent function. Therefore, $\mathbb{E}[(z_{i}^{j})^{2}]$ will come close to $1$, but it will never reach $1$. In addition, $\mathbb{E}[(z_{i}^{j})^{2}]$ will only be $0$ if $c_{i}^{j}=0$. Therefore, $0\leq\mathbb{E}[(z_{i}^{j})^{2}]<1$.
\doubleEnter
Second, consider the value $\alpha^{j}$. In simulations, it is assumed that $0<\alpha^{j}<\infty$, for which $c_{i}^{j}$ is always well-defined. However, for completeness, one can look at the boundary values $\alpha^{j}=0$ and $\alpha^{j}=\infty$ by considering the behavior of $c_{i}^{j}$ in the limits $\alpha^{j}\downarrow0$ and $\alpha^{j}\to\infty$. Let $0\leq\mathbb{E}[(z_{i}^{j})^{2}]<1$ be constant. Then, $c_{i}^{j}$ in the limiting values of $\alpha^{j}\downarrow0$ and $\alpha^{j}\to\infty$ is given by 
\begin{equation}
    \lim_{\alpha^{j}\downarrow0}c_{i}^{j}=\lim_{\alpha^{j}\downarrow0}\frac{\mathbb{E}[(z_{i}^{j})^{2}]}{\mathbb{E}[(z_{i}^{j})^{2}]+(\alpha^{j})^{-2}}=0\quad\text{and}\quad\lim_{\alpha^{j}\to\infty}c_{i}^{j}=\lim_{\alpha^{j}\to\infty}\frac{\mathbb{E}[(z_{i}^{j})^{2}]}{\mathbb{E}[(z_{i}^{j})^{2}]+(\alpha^{j})^{-2}}=1.
\end{equation}
There is an intuitive relation between $\alpha^{j}$ and $\mathbb{E}[(z_{i}^{j})^{2}]$, which is clearer if Equation \ref{eq:theory_diagonal_conceptors_ci} is written as
\begin{equation}
\label{eq:ci_equation}
    c_{i}^{j}=\frac{1}{1+\frac{1}{(\alpha^{j})^{2}\mathbb{E}[(z_{i}^{j})^{2}]}}.
\end{equation}
Notice that excited neurons, i.e. $\mathbb{E}[(z_{i}^{j})^{2}]$ close to $1$, yield larger values of $c_{i}^{j}$, hence those neuron states remain mostly untouched. Conversely, less excited neurons, i.e., $\mathbb{E}[(z_{i}^{j})^{2}]$ close to $0$, yield smaller values of $c_{i}^{j}$, hence those neuron states are suppressed. How much the neurons are untouched or suppressed depends on $\alpha^{j}$. Increasing $\alpha^{j}$ also increases $c_{i}^{j}$ and, conversely, decreasing $\alpha^{j}$ also decreases $c_{i}^{j}$. However, this behavior should not come as a surprise, as this is exactly how the loss function was designed.

\subsubsection{Morphing}
\label{sec:diagonal_conceptors_morphing}
Morphing conceptors can be directly translated to diagonal conceptors. Let the reservoir be loaded with patterns $j_{1}$ and $j_{2}$ and let the associated diagonal conceptors $D^{j_{1}}$ and $D^{j_{2}}$ be given. Then, similar to morphing conceptors, a mixture of the reservoir dynamics associated with patterns $j_{1}$ and $j_{2}$ is obtained by taking a linear combination of $D^{j_{1}}$ and $D^{j_{2}}$ as follows:
\begin{equation}
\label{eq:diagonal_conceptors_morphing}
\begin{aligned}
    \mathbf{r}(n+1)&=\tanh{(W\mathbf{z}(n)+\mathbf{b})},\\
    \mathbf{z}(n+1)&=\big((1-\mu)D^{j_{1}} + \mu D^{j_{2}}\big)\mathbf{r}(n+1),
\end{aligned}
\end{equation}
where $\mu\in\mathbb{R}$ is the mixture parameter. As mentioned in Section \ref{sec:conceptors_morphing}, the mixture parameter for morphing conceptors could be used to not only interpolate between reservoir states, but also extrapolate. However, in the examples shown in Section \ref{sec:simulations_morphing}, interpolation was possible, but extrapolation was not.
\doubleEnter
Similar to morphing with conceptors, the morph is not restricted to two patterns. Again, in the more general case, a morph could be performed for $p$ patterns, where the term $(1-\mu)D^{j_{1}} + \mu D^{j_{2}}$ would be substituted for $\sum_{i=1}^{p}\mu_{i}D^{j_{i}}$. The morphing simulations in this report only make use of the case where $p=2$. 

\subsubsection{Diagonal autoconceptors}
\label{sec:diagonal_autoconceptors}
The arguments for autoconceptors can be translated directly to diagonal conceptors, creating diagonal autoconceptors. If the diagonal conceptor matrix $D$ is made time-dependent, it yields the following update equations:
\begin{equation}
\label{eq:diagonal_autoconceptors_update_equation}
\begin{aligned}
    \mathbf{r}(n+1)&=\tanh{(W\mathbf{z}(n)+\mathbf{b})},\\
    \mathbf{z}(n+1)&=D(n)\mathbf{r}(n+1).
\end{aligned}
\end{equation}
The adaptation rule for $D(n)=\text{diag}(\mathbf{c}(n))$, where $\mathbf{c}(n)$ is the time-dependent conception vector, can be written element-wise, and is read directly from Equation \ref{eq:L_ci_derivative}. This yields
\begin{equation}
\label{eq:diagonal_autoconceptors_adaptation_rule}
    c_{i}(n+1) = c_{i}(n) + \lambda_{i}\Big( \big(1-c_{i}(n)\big)z_{i}^{2}(n) - \alpha^{-2}c_{i}(n)\Big),
\end{equation}
where $c_{i}(n)$ and $z_{i}(n)$ are the $i$-th elements of $\mathbf{c}(n)$ and $\mathbf{z}(n)$, respectively, and $\lambda_{i}$ is the learning rate associated with neuron $i$. Note that the learning rate can be set individually for each neuron, hence lifting the constraint of a global learning rate for autoconceptors. Furthermore, the biological implausibility of autoconceptors is lifted as well, since each neuron state is updated independently of each other. Therefore, in contrast to conceptors, all the information required for updating a conception weight at a synapse is available at that synapse. This does not directly imply biological plausibility, but it certainly does not deny it. This is argued in more detail in Section 3.15 of the conceptors report and this argumentation can be used for diagonal autoconceptors as well \cite{MonsterReport}, so for more information the reader is referred there.
\begin{figure}[t]
    \centering
    \includegraphics[width=\textwidth]{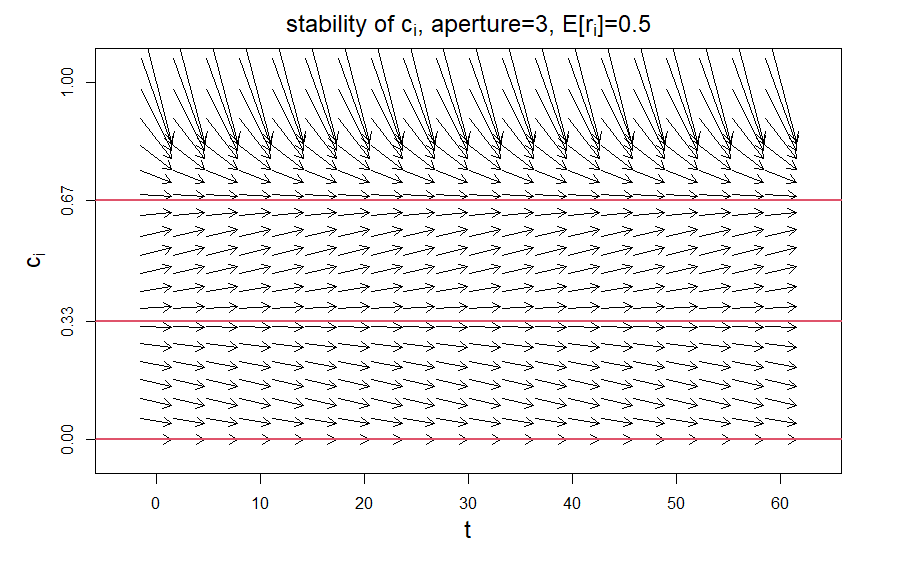}
    \caption{A direction field of Equation \ref{eq:ci_differential_equation}. The solutions, as computed in Equation \ref{eq:ci_stability_solution}, are plotted as red lines.}
    \label{fig:stability_ci}
\end{figure}
\doubleEnter
Furthermore, the dynamics of diagonal autoconceptors are easier to analyse than the dynamics of autoconceptors, because it comprises scalars instead of matrices. A short, intuitive analysis is given here. It will only grasp a small portion of the full dynamics of the reservoir, but it will give some insight nonetheless. Writing the adaptation rule given in Equation \ref{eq:diagonal_autoconceptors_adaptation_rule} as a continuous-time differential equation and letting $z_{i}=c_{i}r_{i}$ gives
\begin{equation}
\label{eq:ci_differential_equation}
    \Dot{c_{i}}(t) = \big(1-c_{i}(t)\big)c_{i}^{2}(t)r_{i}^{2}(t) - \alpha^{-2}c_{i}(t).
\end{equation}
Setting this equal to $0$ and solving for $c_{i}(t)$ gives
\begin{equation}
\label{eq:ci_stability_solution}
    c_{i,\pm}=\frac{1}{2}\pm\frac{1}{2}\sqrt{1-4\alpha^{-2}(\mathbb{E}[r_{i}^{2}])^{-1}}\quad\text{and}\quad c_{i}=0.
\end{equation}
This is a special case of what is more generally shown in Section 3.14.4 of the conceptors report \cite{MonsterReport}. Upon closer inspection it turns out that only $c_{i,+}$ and $c_{i}=0$ are stable solutions, which is neatly shown in Figure \ref{fig:stability_ci}.
\doubleEnter
Figure \ref{fig:stability_ci} shows that over time the value of $c_{i}$ will either go towards the upper red line, corresponding to $c_{i,+}$ in Equation \ref{eq:ci_stability_solution}, or $c_{i}=0$. In can also be seen that $c_{i,-}$ is not a stable solution. Furthermore, the curvature is much larger for values above $c_{i,+}$ than for values $c_{i,+}$, hence the convergence rate in those areas will most likely differ. Therefore, in simulations one could, for example, increase the learning rate for values of $c_{i}$ smaller than $0.5$ to increase the rate of convergence.
\doubleEnter
At the present moment, diagonal autoconceptors remain mostly unexplored. A more comprehensive analysis is required to fully understand the dynamics of this system.

\newpage
\section{Simulation Methods}
\label{sec:simulation_methods}
Several simulations will be discussed in Section \ref{sec:simulations} to demonstrate the capabilities of diagonal conceptors compared to conceptors. This section describes how those simulations were set up. In addition, this section describes how to train conceptors and diagonal conceptors. The last part of this section describes how the accuracy of the conceptors and diagonal conceptors is determined.

\subsection{Initializing the Parameters}
\label{sec:initializing_the_parameters}
When setting up computer simulations, the RNN must be initialized appropriately. How many connections between reservoir neurons are optimal and how should they be weighted? From what distribution should they be drawn? The same questions can be asked for the input weights and the bias vector. This section introduces some of the rules of thumb for initializing the simulation parameters.
\doubleEnter
Remember that the system is given by the equations
\begin{equation}
\label{eq:training_conceptors_update_equations1}
    \mathbf{x}^{j}(n+1) = \tanh{(W^{*}\mathbf{x}^{j}(n) + W^{in}\mathbf{p}^{j}(n+1) + \mathbf{b})}
\end{equation}
where $\mathbf{x}^{j}(n)\in\mathbb{R}^{N}$ is the reservoir state associated with pattern $j$ at time step $n$, $\mathbf{p}^{j}(n)\in\mathbb{R}^{M}$ is the $n$-th time step of pattern $j$, $\mathbf{b}\in\mathbb{R}^{N}$ is the bias vector, $W^{*}\in\mathbb{R}^{N\times N}$ are the reservoir weights, and $W^{in}\in\mathbb{R}^{N\times M}$. The simulations in this report will use a slightly different system consisting of so-called \textit{leaky-integrated} neurons, which changes Equation \ref{eq:training_conceptors_update_equations1} to
\begin{equation}
\label{eq:training_conceptors_update_equations2}
\begin{aligned}
    \mathbf{\Tilde{x}}^{j}(n+1) &= \tanh{(W^{*}\mathbf{x}^{j}(n) + W^{in}\mathbf{p}^{j}(n+1) + \mathbf{b})}\\
    \mathbf{x}^{j}(n+1)&=(1-\alpha_{l})\mathbf{x}^{j}(n)+\alpha_{l}\mathbf{\Tilde{x}}^{j}(n+1)
\end{aligned}
\end{equation}
where $0\leq\alpha_{l}\leq1$ is the leaking rate. Leaky-integrated neurons are an extension of the non-leaky integrated neurons that were described so far, since Equation \ref{eq:training_conceptors_update_equations1} is recovered for $\alpha_{l}=1$, and they offer better fine-tuning of the reservoir dynamics \cite{LeakyNeurons}. The leaking rate is a so-called \textit{exponential smoother} and it determines how much of the previous state is still remembered in computing the next state. Adding the leaking rate into the update equation does not change anything that has been discussed up until this point. Intuitively, lowering the leaking rate results in a smoother output pattern, since big deviations are slightly suppressed.
\doubleEnter
The elements of $W^{in}$, $\mathbf{b}$ and $W^{*}$ are all drawn from the standard normal distribution and scaled appropriately. The input weights matrix and bias vector are full matrices, whereas the reservoir weights matrix is a sparse matrix. The sparsity of the reservoir weights matrix is determined by the number of neurons and can be chosen in different manners \cite{PracticalGuideToApplyingEchoStateNetworks}. For simulations discussed in this report, the sparsity was set to $1$ if $N<20$ and $10/N$ if $N\geq20$, where $N$ is the number of reservoir neurons.
\doubleEnter
The scaling of $W^{in}$ and $\mathbf{b}$ mostly determines how much the neurons in the reservoir are excited. For example, lower scalings of $W^{in}$ and $\mathbf{b}$ result in less excited states. Intuitively, if the neurons are too excited their values will be close to $-1$ or $1$, which will make the neurons act more in a binary switching way. This is not ideal for smooth patterns. On the other hand, if the neurons are not excited enough, their values will be close to $0$, which is desired for very linear tasks, but not for non-linear tasks.
\doubleEnter
Furthermore, the scalings of $W^{in}$, $\mathbf{b}$, and the spectral radius of $W^{*}$ must be appropriate to ensure that the RNN possesses the so-called \textit{echo state property} \cite{EchoStateProperty}. The echo state property ensures that the RNN "forgets" the initial state. Often, the initial state of a network is not known and will simply be set to the all-zero vector. Therefore, the network must be given some time to \textit{washout} the initial state of the system. This is ensured via a washout period in the simulation. However, the initial state of the RNN must also be forgotten and the echo state property makes sure this happens. The length of the washout period, denoted by $n_{washout}$, can be found by driving the reservoir from two initial states, e.g. the all-zero vector and the all-ones vector, and then determining how long it takes for the two reservoir states to converge, up to a residual error, to the same state. Note that $n_{washout}$ depends on the other parameters of the simulation, e.g., a lower leaking rate will require a longer washout period.
\doubleEnter
Setting the parameters mentioned above is mainly a matter of trial and error. In order to get some intuition into the parameters, it is advised to manually adjust a single parameter and see how it changes the dynamics of the reservoir or the outputs. It is possible to implement parameter optimization algorithms, but when the output requires visual inspection it may be better to try manually adjusting the parameters one-by-one. Much research has been conducted in finding good scaling parameters, which has been concisely summarized in the paper by M. Lukoševičius, \textit{A Practical Guide to Applying Echo State Networks} \cite{PracticalGuideToApplyingEchoStateNetworks}. For more intuition on setting up the parameters, the reader is referred to that paper.

\subsection{Training Conceptors}
\label{sec:training_conceptors}
This section describes how conceptors are trained. Let there be a set of patterns $\mathbf{P}=\{\mathbf{p}^{1},\mathbf{p}^{2},...,\mathbf{p}^{p}\}$, where $\mathbf{p}^{j}\in\mathbb{R}^{M}$ for all $j$. Initialize the following:
\begin{itemize}
    \item number of neurons in the reservoir ($N$),
    \item input weights ($W^{in}\in\mathbb{R}^{N\times M}$): Random values drawn from the standard normal distribution and scaled appropriately,
    \item bias ($\mathbf{b}\in\mathbb{R}^{N}$): Random values drawn from the standard normal distribution and scaled appropriately,
    \item initial reservoir weights ($W^{*}\in\mathbb{R}^{N\times N}$): Sparse matrix with random values drawn from the standard normal distribution and scaled appropriately,
    \item leaking rate $\alpha_{l}$.
\end{itemize}
\doubleEnter
\textbf{Washout}\\
Drive the reservoir with pattern $j$ for $n_{washout}$ time steps according to Equation \ref{eq:training_conceptors_update_equations2}. All information during this period is discarded. The initial state can be initialized as any random vector in $[-1,1]^{N}$, but usually $\mathbf{x}^{j}(0)=\mathbf{0}$, for all $j$, for simplicity.
\doubleEnter
\textbf{Learning}\\
The reservoir is continued to be driven for another $n_{learn}$ time steps. During the learning period, the aim is to compute:
\begin{itemize}
    \item $R^{j}=X^{j}(X^{j})^{T}/n_{learn}$, where $X^{j}$ is the state collection matrix,
    \item $W$, such that $W\mathbf{x}^{j}(n)\approx W^{*}\mathbf{x}^{j}(n)+W^{in}\mathbf{p}^{j}(n)$ for all $n,j$,
    \item $W^{out}$, such that $W^{out}\mathbf{x}^{j}(n)\approx\mathbf{p}^{j}(n)$ for all $n,j$,
\end{itemize}
so the following vectors are collected:
\begin{itemize}
    \item $\mathbf{x}^{j}(n)$, to create: $X^{j}$, where $(X^{j})_{*,n}=\mathbf{x}^{j}(n_{washout}+n)$,
    \item $\mathbf{x}^{j}(n-1)$, to create: $\Tilde{X}^{j}$, where $(\Tilde{X}^{j})_{*,n}=\mathbf{x}^{j}(n_{washout}+n-1)$,
    \item $\mathbf{p}^{j}(n)$, to create: $P^{j}$, where $(P^{j})_{*,n}=\mathbf{p}^{j}(n_{washout}+n)$
    \item $W^{*}\mathbf{x}^{j}(n)+W^{in}\mathbf{p}^{j}(n)$, to create: $W_{target}^{j}$, where\\ $(W_{target}^{j})_{*,n}=W^{*}\mathbf{x}^{j}(n_{washout}+n)+W^{in}\mathbf{p}^{j}(n_{washout}+n)$.
\end{itemize}
In the above, $(.)_{*,n}$ is used to denote the $n$-th column of a matrix.

\doubleEnter
\textbf{Computing Conceptors}\\
The washout and learning period is repeated for all patterns $j=1,2,...,P$, during which the state correlation matrices $R^{j}$ are collected. Furthermore, the aperture $\alpha^{j}$ is set for each pattern $j$. The conceptor matrix $C^{j}$ is computed with  
\begin{equation}
    C^{j}=R^{j}(R^{j}+(\alpha^{j})^{-2}I)^{-1}.
\end{equation}
The aperture is chosen heuristically, but there exist optimization techniques, see Section 3.8 in \cite{MonsterReport}. These techniques will not be employed in this report.
\doubleEnter
\textbf{Computing $W$ and $W^{out}$}\\
The collected matrices $X^{j}$, $\Tilde{X}^{j}$, $P^{j}$, and $W^{j}_{target}$ are concatenated, which gives $X=[X^{1}|X^{2}|...|X^{p}]$, $\Tilde{X}=[\Tilde{X}^{1}|\Tilde{X^{2}}|...|\Tilde{X}^{p}]$, $P=[P^{1}|P^{2}|...|P^{p}]$, and $W_{target}=[W_{target}^{1}|W_{target}^{2}|...|W_{target}^{p}]$. Following Section \ref{sec:theory_storing_patterns}, the patterns are stored and the output weights are computed via ridge regression as follows:
\begin{equation}
    W = (\Tilde{X}\Tilde{X}^{T}+\varrho^{W}I)^{-1}\Tilde{Z}W_{target}^{T},\quad W^{out} = (XX^{T}+\varrho^{W^{out}}I)^{-1}XP^{T},
\end{equation}
where $\varrho^{W}$ and $\varrho^{W^{out}}$ are the regularization constants for $W$ and $W^{out}$ respectively, and $I$ is the identity matrix. The regularization constants have a big influence on the output of the simulation. A small regularization constant may result in overfitting, whereas a large regularization constant may lead to underfitting. The regularization constants are essentially another two parameters in the simulation and they need to be chosen appropriately. Again, this is done heuristically. 
\doubleEnter
\textbf{Self-generating reservoir}\\
After $W$, $W^{out}$, and $C^{j}$ have been computed, the reservoir should be able to self-generate any pattern $j$ by inserting $C^{j}$ in the update equation, yielding
\begin{equation}
\begin{aligned}
    \mathbf{r}(n+1)&=(1-\alpha_{l})\mathbf{r}(n)+\alpha_{l}\tanh{(W\mathbf{z}(n) + \mathbf{b})}\\
    \mathbf{z}(n+1)&=C^{j}\mathbf{r}(n+1)
\end{aligned}
\end{equation}
and 
\begin{equation}
    \mathbf{y}(n)=W^{out}\mathbf{z}(n),
\end{equation}
where $\mathbf{y}$ is expected to behave like pattern $j$.

\subsection{Training Diagonal Conceptors} \label{sec:training_diagonal_conceptors}
The training scheme of diagonal conceptors was inspired by the training scheme of RFC, described in Section 3.15 of the conceptors report \cite{MonsterReport}. The main contrast between the two training schemes is that in the case of diagonal conceptors, the conception weights are initialized to randomly drawn values from the uniform distribution on $[0,1]$ instead of $1$. This random initialization is an important feature of the training scheme of diagonal conceptors and it requires more attention than can be given in this section. Therefore, the consequences of the random initialization will be discussed in more detail in Section \ref{sec:discussion_initial_diagonal_conceptors}.
\doubleEnter
Furthermore, the learning period of the training scheme of diagonal conceptors comprises two stages. In stage 1 of the learning period, hereafter referred to as \textit{stage 1}, the diagonal conceptors are computed. In stage 2 of the learning period, hereafter referred to as \textit{stage 2}, the patterns are stored in the reservoir.
\doubleEnter
The training commences with initializing the following:
\begin{itemize}
    \item number of neurons in reservoir ($N$),
    \item input weights ($W^{in}\in\mathbb{R}^{N\times M}$): Random values drawn from the standard normal distribution and scaled appropriately,
    \item bias ($\mathbf{b}\in\mathbb{R}^{N}$): Random values drawn from the standard normal distribution and scaled appropriately,
    \item initial reservoir weights ($W^{*}\in\mathbb{R}^{N\times N}$): Sparse matrix with random values drawn from the standard normal distribution and scaled appropriately,
    \item leaking rate $\alpha_{l}$,
    \item initial diagonal conceptors ($D_{0}^{j}=\text{diag}(\mathbf{c_{0}}^{j})$, where $\mathbf{c_{0}}^{j}\in[0,1]^{N}$): Elements of $\mathbf{c_{0}}^{j}$ are randomly drawn from the uniform distribution on the interval $[0,1]$,
\end{itemize}
\doubleEnter
\textbf{Washout}\\
The reservoir is driven with pattern $j$ for $n_{washout}$ time steps according to the following update equations:
\begin{equation}
\label{eq:training_diagonal_conceptors_update_equations}
\begin{aligned}
    \mathbf{r}^{j}(n+1)&=(1-\alpha_{l})\mathbf{r}^{j}(n)+\alpha_{l}\tanh{(W^{*}\mathbf{z}^{j}(n) + W^{in}\mathbf{p}^{j}(n+1) + \mathbf{b})},\\
    \mathbf{z}^{j}(n+1)&=D_{0}^{j}\mathbf{r}(n+1),
\end{aligned}
\end{equation}
where the initial state $\mathbf{z}^{j}(0)$ is, again, simply set to the all-zero vector.
\doubleEnter
\textbf{Learning - Stage 1}\\
After the washout period, the reservoir is continued to be driven for $n_{stage1}$ time steps according to Equation \ref{eq:training_diagonal_conceptors_update_equations}, but now the states $\mathbf{z}^{j}(n)$ are collected in the state collection matrix
\begin{align*}
    Z^{j}=
    \begin{bmatrix}
    \mathbf{z}^{j}(n_{washout}+1) & \mathbf{z}^{j}(n_{washout}+2) &
    \hdots &
    \mathbf{z}^{j}(n_{washout}+n_{stage1})
    \end{bmatrix}\in\mathbb{R}^{N\times n_{stage1}}.
\end{align*}
As shown in Section \ref{sec:theory_diagonal_conceptors}, the conception weights $c_{i}^{j}$ can be computed element-wise
\begin{equation}
    c^{j}_{i}=\frac{\mathbb{E}[(z^{j}_{i})^{2}]}{\mathbb{E}[(z^{j}_{i})^{2}]+(\alpha^{j})^{-2}},
\end{equation}
where $z^{j}_{i}$ is the $i$-th element of $\mathbf{z}^{j}$ and $\alpha^{j}$ is the aperture for pattern $j$. Note that in simulations, $\mathbf{c}^{j}$ is computed in a vectorized manner, which gives 
\begin{equation}
    \mathbf{c}^{j}=\mathbb{E}[\mathbf{z}^{j}\circ\mathbf{z}^{j}]\oslash(\mathbb{E}[\mathbf{z}^{j}\circ\mathbf{z}^{j}]+(\alpha^{j})^{-2}),
\end{equation}
where $\circ$ is the Hadamard product, or the element-wise multiplication operator, and $\oslash$ is the element-wise division operator. Furthermore, $\mathbb{E}[\mathbf{z}^{j}\circ\mathbf{z}^{j}]$ is approximated by
\begin{equation}
    \mathbb{E}[\mathbf{z}^{j}\circ\mathbf{z}^{j}]\approx\frac{1}{n_{stage1}}\sum_{l}^{n_{stage1}}(Z^{j}\circ Z^{j})_{*,l},
\end{equation}
which states that $Z^{j}$ is multiplied with itself element-wise and then the mean of each row is taken. $(Z^{j}\circ Z^{j})_{*,l}$ denotes the $l$-th column of $Z^{j}\circ Z^{j}$.
\doubleEnter
\textbf{Learning - Stage 2}\\
The reservoir is continued to be driven for another $n_{stage2}$ time steps, but now with the newly computed diagonal conceptors $D^{j}=\text{diag}(\mathbf{c}^{j})$ in the loop rather than $D_{0}^{j}$. This gives the new, slightly different update equations:
\begin{equation}
\label{eq:simulation_methods_training_diagonal_conceptors_update_equations2}
\begin{aligned}
    \mathbf{r}^{j}(n+1)&=(1-\alpha_{l})\mathbf{r}^{j}(n)+\alpha_{l}\tanh{(W^{*}\mathbf{z}^{j}(n) + W^{in}\mathbf{p}^{j}(n+1) + \mathbf{b})},\\
    \mathbf{z}^{j}(n+1)&=D^{j}\mathbf{r}(n+1).
\end{aligned}
\end{equation}
During stage 2, the goal is to compute:
\begin{itemize}
    \item $W$, such that $W\mathbf{z}^{j}(n)\approx W^{*}\mathbf{z}^{j}(n)+W^{in}\mathbf{p}^{j}(n)$ for all $n,j$,
    \item $W^{out}$, such that $W^{out}\mathbf{z}^{j}(n)\approx\mathbf{p}^{j}(n)$ for all $n,j$,
\end{itemize}
so the following vectors are collected:
\begin{itemize}
    \item $\mathbf{r}^{j}(n)$, to create: $X^{j}$, where $(X^{j})_{*,n}=\mathbf{r}^{j}(n_{0}+n)$,
    \item $\mathbf{z}^{j}(n-1)$, to create: $\Tilde{Z}^{j}$, where $(\Tilde{Z}^{j})_{*,n}=\mathbf{z}^{j}(n_{0}+n-1)$,
    \item $\mathbf{p}^{j}(n)$, to create: $P^{j}$, where $(P^{j})_{*,n}=\mathbf{p}^{j}(n_{0}+n)$,
    \item $W^{*}\mathbf{z}^{j}(n)+W^{in}\mathbf{p}^{j}(n)$, to create: $W_{target}^{j}$, where\\ $(W_{target}^{j})_{*,n}=W^{*}\mathbf{z}^{j}(n_{0}+n)+W^{in}\mathbf{p}^{j}(n_{0}+n)$,
\end{itemize}
where $n_{0}=n_{washout}+n_{stage1}$ and $(.)_{*,n}$ denotes the $n$-th column of a matrix.
\doubleEnter
\textbf{Computing $W$ and $W^{out}$}\\
The washout, stage 1, and stage 2 are repeated for all patterns $j=1,2,...,p$ and the collected matrices are concatenated. This gives $X=[X^{1}|X^{2}|...|X^{p}]$, $\Tilde{Z}=[\Tilde{Z}^{1}|\Tilde{Z^{2}}|...|\Tilde{Z}^{p}]$, ${P=[P^{1}|P^{2}|...|P^{p}]}$, and $W_{target}=[W_{target}^{1}|W_{target}^{2}|...|W_{target}^{p}]$. Using ridge regression, the new reservoir weights $W$ and the output weights $W^{out}$ are then computed as follows:
\begin{equation}
    W = (\Tilde{Z}\Tilde{Z}^{T}+\varrho^{W}I)^{-1}\Tilde{Z}W_{target}^{T},\quad W^{out} = (XX^{T}+\varrho^{W^{out}}I)^{-1}XP^{T},
\end{equation}
where $\varrho^{W}$ and $\varrho^{W^{out}}$ are the regularization constants for $W$ and $W^{out}$ respectively, and $I$ is the identity matrix.
\doubleEnter
\textbf{Self-generating reservoir}\\
After computing $W$ and $W^{out}$, the reservoir is expected to self-generate pattern $j$ by letting the reservoir run according to the update equations 
\begin{equation}
\begin{aligned}
    \mathbf{r}^{j}(n+1)&=(1-\alpha_{l})\mathbf{r}^{j}(n)+\alpha_{l}\tanh{(W\mathbf{z}^{j}(n) + \mathbf{b})},\\
    \mathbf{z}^{j}(n+1)&=D^{j}\mathbf{r}(n+1),
\end{aligned}
\end{equation}
where the output can be read by
\begin{equation}
    \mathbf{y}(n+1)=W^{out}\mathbf{r}(n).
\end{equation}
The output $\mathbf{y}$ is expected to behave like pattern $j$.

\subsection{Pattern Comparison}
\label{sec:simulation_methods_pattern_comparison}
Let $P=[\mathbf{p}(1)\;\mathbf{p}(2)\;...\;\mathbf{p}(L)]^{T}\in\mathbb{R}^{L\times M}$ and ${Y=[\mathbf{y}(1)\;\mathbf{y}(2)\;...\;\mathbf{y}(L')]^{T}\in\mathbb{R}^{L'\times M}}$ denote the input pattern and the pattern that is self-generated by the reservoir, respectively. When comparing $Y$ and $P$, the first step is always visual inspection. Especially when choosing the simulation parameters, it is useful to see how the self-generated pattern compares to the input pattern. The second step is quantifying the comparison. This is practical in case the patterns are high-dimensional or many patterns are compared simultaneously. In this report, the accuracy of the match between a self-generated pattern and an input pattern will be compared with the Normalized Root Mean Square Error (NRMSE).
\doubleEnter
Let $\mathbf{p}_{obs}\in\mathbb{R}^{K}$ be an \textit{observed} pattern and let $\mathbf{p}_{target}\in\mathbb{R}^{K}$ be a \textit{target} pattern. The NRMSE of $\mathbf{p}_{obs}$ and $\mathbf{p}_{target}$ is given by
\begin{equation}
    NRMSE(\mathbf{p}_{obs},\mathbf{p}_{target})= \sqrt{\frac{\frac{1}{K}\sum\limits_{n=1}^{K}(\mathbf{p}_{obs}(n)-\mathbf{p}_{target}(n))^{2}}{\text{Var}(\mathbf{p}_{target})}},
\end{equation}
where $\text{Var}(\mathbf{p}_{target})$ is the variance of $\mathbf{p}_{target}$. The NRMSE is a normalized version of the Root Mean Square Error (RMSE): 
\begin{equation}
    RMSE(\mathbf{p}_{obs},\mathbf{p}_{target})=\sqrt{\frac{1}{K}\sum\limits_{n=1}^{K}(\mathbf{p}_{obs}(n)-\mathbf{p}_{target}(n))^{2}}.
\end{equation}
Note that $\text{NRMSE}=1$ if $\mathbf{p}_{obs}(n)$ is equal to the mean of $\mathbf{p}_{target}$ for all $n$. This indicates that a reasonable model should achieve a NRMSE of between $0$ and $1$. In this report, a simulation has achieved good accuracy if the NRMSE is $0.1$ or lower.
\doubleEnter
If the patterns are higher-dimensional, so $P_{obs}=[\mathbf{p}_{obs}(1)\;\mathbf{p}_{obs}(2)\;...\;\mathbf{p}_{obs}(K)]^{T}\in\mathbb{R}^{K\times M}$ and $P_{target}=[\mathbf{p}_{target}(1)\;\mathbf{p}_{target}(2)\;...\;\mathbf{p}_{target}(K)]^{T}\in\mathbb{R}^{K\times M}$, where $M>1$, then the NRMSE is taken for each column. This yields an $M$-dimensional vector of NRMSEs. One could simply take the mean over all values. However, if the observed pattern has a small NRMSE in one column, but a large NRMSE in another column, this information would get lost in the mean. Therefore, when comparing higher-dimensional patterns, it is more informative to not only take the mean, but also show the smallest and largest NRMSE values, and the standard deviation among all NRMSE values. In this case, a simulation has achieved good accuracy if the mean over all NRMSEs is $0.1$ or lower and the standard deviation is $0.1$ or lower.
\doubleEnter
The length of the input pattern $P$, denoted by $L$, is usually given and cannot be easily altered, unless the data is synthetically generated. The length of the self-generated pattern $Y$, denoted by $L'$, on the other hand, is usually regulated by the simulation and can easily be adjusted. Therefore, it is assumed that $L'\geq L$, where, in the simulations in this report, $L'$ is simply set equal to $L$. In order to compare the input pattern and the self-generated pattern, the comparison length, denoted by $K$, must be chosen appropriately.
\doubleEnter
There is a nontrivial problem to measuring the accuracy between an input pattern and a self-generated pattern. If the observed pattern is matched against the target pattern for a long time, i.e., $K$ is large, small phase shifts can lead to large NRMSEs. For example, if the target pattern is a sine wave with period $1$ and the observed pattern is a sine wave with period $1.001$. Then, after some time, the two patterns will have zero correlation and the NRMSE will be $1$. Consequently, the NRMSE is more reliable if the patterns are compared for a short time, i.e., $K$ is small. However, in that case, one cannot assess whether the generating RNN is long-term stable. Therefore, in this report, the long-term stability is assessed by computing the NRMSE for the first $K$ time steps and the last $K$ time steps. If the NRMSE in both cases is below $0.1$, the match between the input pattern and the self-generated pattern is deemed good and long-term stable.
\doubleEnter
Lastly, the autonomous run of the reservoir is usually started from a random initial state, so the observed pattern will most likely not be phase-aligned with the target pattern. The observed pattern $\mathbf{p}_{obs}$ and the target pattern $\mathbf{p}_{target}$ are phase-aligned by finding a shifting parameter $d\in\mathbb{N}$, with $d\geq1$, such that
\begin{equation}
    NRMSE\Big(\mathbf{p}_{obs}[d:K],\mathbf{p}_{target}[1:(K-d+1)]\Big)
\end{equation}
is minimum, where $\mathbf{p}_{obs}[d:K]$ denotes the subset of $\mathbf{p}_{obs}$ containing only the elements from time step $d$ up to and including $K$, which is the same notation used for indexing in the $R$ programming language. The upper bound of $d$ is set larger than the period of the target pattern to ensure optimal phase alignment. After the optimal $d$ has been found, the shifted patterns are stored as well as their NRMSE. Note that in this setup, the NRMSE is computed over $K-d+1$ time steps. This phase alignment is only performed for $1$-dimensional patterns in this report.

\newpage
\section{Simulations}
\label{sec:simulations}
As mentioned in the previous sections, conceptors were first introduced in the conceptor report \cite{MonsterReport}. In that report, several examples with different patterns are used to highlight the numerous qualities of conceptors. It is for that reason that the patterns used in this section are taken from those examples. This allows for a direct comparison between conceptors and diagonal conceptors. Therefore, this section comprises three examples. A simulation with conceptors and a simulation with diagonal conceptors was conducted for each example. The findings are reported in the first three parts of this section. In addition, a morphing simulation with conceptors as well as a morphing simulation with diagonal conceptors was conducted. This is reported in the last part of this section. Simulation findings will be discussed in Section \ref{sec:discussion}.
\doubleEnter
The patterns used in the three examples increase in difficulty. The first example consists of four periodic patterns, two irrational-periodic sine waves and two $5$-periodic random points. This example serves as a introductory demonstration, which also shows how similar patterns can be distinguished by both conceptors and diagonal conceptors. The second example comprises four patterns sampled from the well-known chaotic attractors, the Lorenz, Rössler, Mackey-Glass, and Hénon attractor. Chaotic attractors are intrinsically unpredictable, which makes them a fitting challenge for conceptors as well as diagonal conceptors. In this example, the change in dynamics due to the aperture is highlighted. The last example models a collection of human motions and it is the most difficult out of the three examples. It comprises $15$ $61$-dimensional patterns of lengths between roughly $200$ and $900$ time steps.
\doubleEnter
All the simulations in this section were implemented in R and the source code can be found \href{https://github.com/jorispdejong/DiagonalConceptors}{here} \cite{GithubRepoDiagonalConceptors}.

\subsection{Periodic Patterns}
\label{sec:simulations_periodic_patterns}
A function $f$ is periodic if $f(t+k)=f(t)$ for some constant $k$. For discrete-time patterns this means that the pattern is either \textit{integer-periodic} or \textit{irrational-periodic}. If a pattern is integer-periodic, there exists a positive integer $k$ such that $\mathbf{p}(n+k)=\mathbf{p}(n)$. If a pattern is irrational-periodic, the pattern is sampled with a sample interval $d\in\mathbb{Z}_{>0}$ from a continuous-time signal with periodicity $k\in\mathbb{R}_{>0}$ and the ratio $d/k$ is irrational. An integer-periodic pattern with period $k$ will occupy $k$ points in state space, as it will simply re-visit those points after every cycle. An irrational-periodic pattern, on the other hand, will occupy a more complex set of points in state space. For example, a one-dimensional irrational-periodic pattern will occupy a set of points that can be characterized topologically as a one-dimensional cycle that is homeomorphic to the unit cycle in $\mathbb{R}^{2}$\cite{MonsterReport}.
\doubleEnter
Let $\mathbf{p}^{1}$ and $\mathbf{p}^{2}$ be sampled from a sine wave with period $\approx8.83$ and $\approx9.83$, respectively. They are irrational-periodic. Let $\mathbf{p}^{3}$ be a random $5$-periodic pattern and let $\mathbf{p}^{4}$ be a $5$-periodic pattern that is a slight variation of $\mathbf{p}^{3}$. The set of patterns is given by $\mathbf{P}=\{\mathbf{p}^{1}, \mathbf{p}^{2}, \mathbf{p}^{3}, \mathbf{p}^{4}\}$. The patterns are $1$-dimensional and are initialized to a length of $L=5000$ time steps. An interval of $30$ time steps is shown in the left column of Figure \ref{fig:c_pp_results}.
\doubleEnter
For both the conceptors and diagonal conceptors simulation, the reservoir is initialized with the number of reservoir neurons $N=100$, leaking rate $\alpha_{l}=1$, and regularization constants $\varrho^{W}=0.001$ and $\varrho^{W^{out}}=0$. The reservoir weights $W^{*}$ and the input weights $W^{in}$ are both scaled by $1$ and the bias $\mathbf{b}$ is scaled by $0.2$. The length of the washout period is $n_{washout}=200$. In the case of conceptors, the apertures was set to $\alpha^{j}=100$ and for diagonal conceptors it was set to $\alpha^{j}=8$, for all $j$. The length of the learning period for conceptors was set to $n_{learn}=L-n_{washout}=4800$ time steps. For the diagonal conceptors, the length of \textit{stage 1} was set to ${n_{stage1}=500}$, and the length of stage 2 was set to $n_{stage2}=L-n_{washout}-n_{stage1}=4400$ time steps. For comparability, the network size $N$ is set equal to the the network size in the example of the conceptors report \cite{MonsterReport}.
\doubleEnter
The goal of this example is to demonstrate that diagonal conceptors can distinguish similar patterns with just as little effort as conceptors. 

\subsubsection{Conceptors}
\label{sec:simulations_pp_c}
\begin{figure}[ht]
    \centering
    \includegraphics[width=\textwidth]{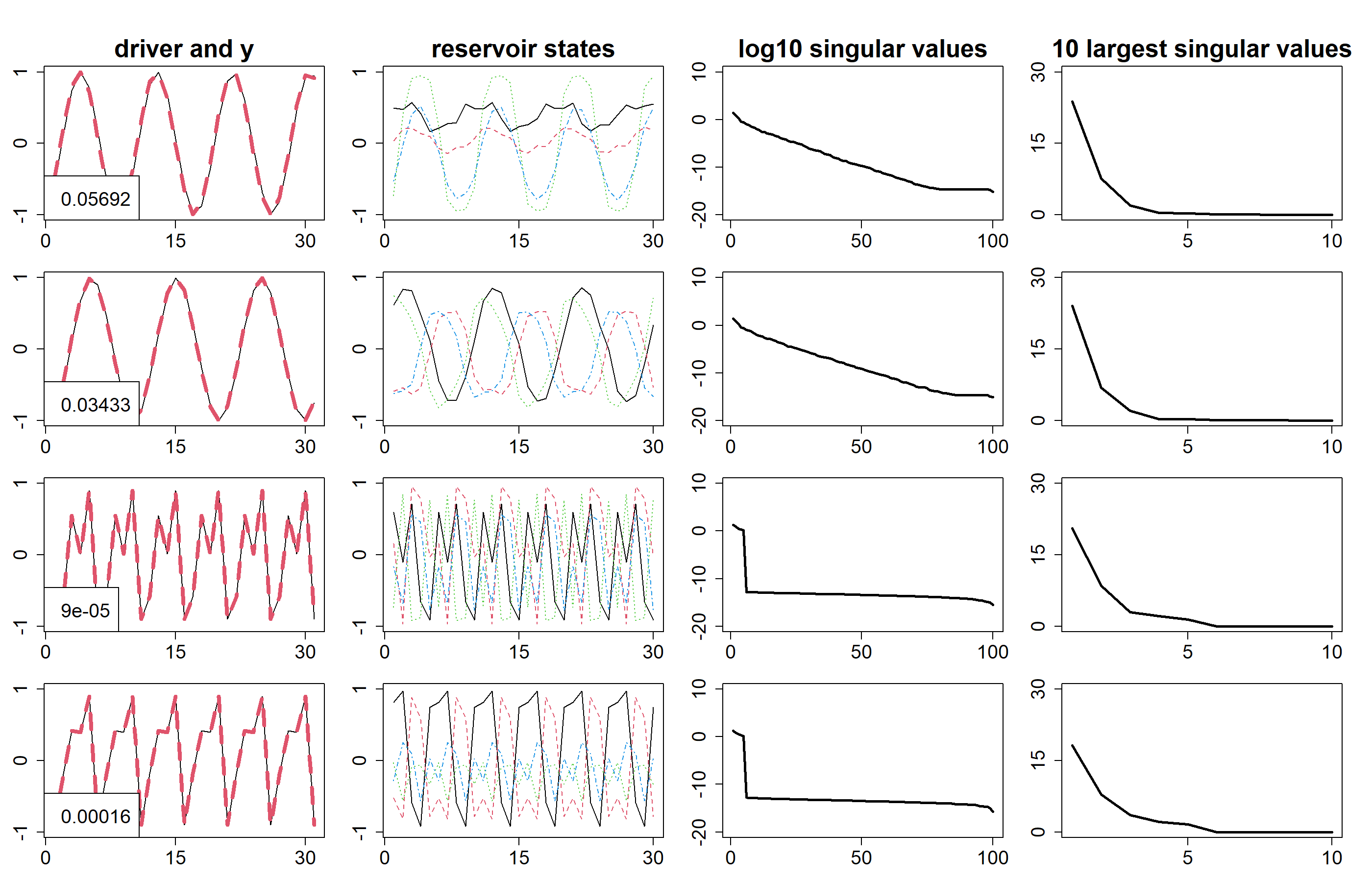}
    \caption{Row $j$ depicts the results for pattern $j$. The left column shows the first $30$ time steps of the driving patterns (black lines) and the conceptor-generated patterns (red dashed lines) and the NRMSE between the patterns over all time steps $n$. The second column shows the first $30$ time steps of $4$ randomly chosen neuron states. The third shows the $\text{log}_{10}$ of the singular values of $R^{j}$ and the fourth column shows the $10$ largest singular values of $R^{j}$.}
    \label{fig:c_pp_results}
\end{figure}

Lets start with the conceptors simulation. The reservoir was driven with the four patterns and the conceptors were trained according to the training algorithm described in Section \ref{sec:training_conceptors}. The conceptor matrices were stored and the reservoir was autonomously run for $n_{learn}=4800$ time steps with $C^{j}$, for all $j$, yielding the self-generated patterns, denoted by $\mathbf{y}^{j}$. The self-generated patterns were phase-aligned with the input patterns, as described in Section \ref{sec:simulation_methods_pattern_comparison}.
\doubleEnter
Both the self-generated patterns and the target patterns are plotted on a small interval, which can be seen in the first column of Figure \ref{fig:c_pp_results}, where the target pattern (driver) is depicted by the black line, the conceptors-generated pattern ($\mathbf{y}$) is depicted by the red dashed line, and the NRMSE is shown in the bottom left corner. In the second column of Figure \ref{fig:c_pp_results}, the neuron state of four randomly chosen neurons, in the same interval as the first column, is plotted. From the similarity between the neuron state and the input pattern as well as the periodicity of the neuron state, it can be inferred that the dynamics of the patterns have been captured by the reservoir.
\doubleEnter
To illustrate the difference in the dynamics of the reservoir between the integer- and irrational-periodic patterns, the $\text{log}_{10}$ of the singular values of $R^{j}$ and the $10$ largest singular values of $R^{j}$ are plotted in the third and fourth columns of Figure \ref{fig:c_pp_results}, respectively. Let $R^{j}=U\Sigma U^{T}$ be the singular value decomposition of $R^{j}$. The singular values of $R^{j}$ are given by the diagonal of the diagonal matrix $\Sigma$ and the principal components of $R^{j}$ are given by $U$. The third column of Figure \ref{fig:c_pp_results} neatly shows that for the irrational-periodic patterns, the principal components of $R^{j}$ span all of $R^{N}$, whereas the principal components of the $5$-periodic patterns are only non-zero in $5$ directions, as the reservoir periodically visits the same $5$ states.
The fourth column shows that a large contribution of the singular values is concentrated on a few principal directions for all four patterns. This means that the point cloud created by the reservoir states, whose characteristics are captured by $R^{j}$, has a negligible variance in most directions. The small singular values for the $5$-periodic patterns are due to numerical errors.
\doubleEnter
Adjusting the simulation parameters may result in slightly better or worse performance. Specifically, the conceptors still achieve good accuracy, as defined in Section \ref{sec:simulation_methods_pattern_comparison}, when the parameters are individually modified in the following ranges:
\begin{itemize}
    \item $W^{in}$ scaling: $[0.6,2]$,
    \item $W^{*}$ scaling: $[0.8,1]$,
    \item $\mathbf{b}$ scaling: $[0.2,0.6]$,
    \item $\alpha_{l}$: $[0.9,1]$,
    \item $\varrho^{W^{out}}$: $[0,1000]$,
    \item $\varrho^{W}$: $[0,0.01]$,
    \item $\alpha^{j}$: $[10,10000]$, for all $j$.
\end{itemize}
It must be noted that these ranges indicate that good accuracy is still achieved if a single parameter is adjusted to anywhere in its given range, while the other parameters remain unchanged. Regardless, there are combinations of parameters possible that lie outside of these ranges, e.g., the following parameters also yield good accuracy: $W^{in}$ scaling $=0.5$, $W^{*}$ scaling $=1$, $\mathbf{b}$ scaling $=0.7$, $\alpha_{l}=1$, $\varrho^{W^{out}}=0.000001$, $\varrho^{W}=0.001$, $\alpha^{j}=100$ for all $j$.

\subsubsection{Diagonal Conceptors}
\label{sec:simulations_pp_dc}
\begin{figure}[ht]
    \centering
    \includegraphics[width=\textwidth]{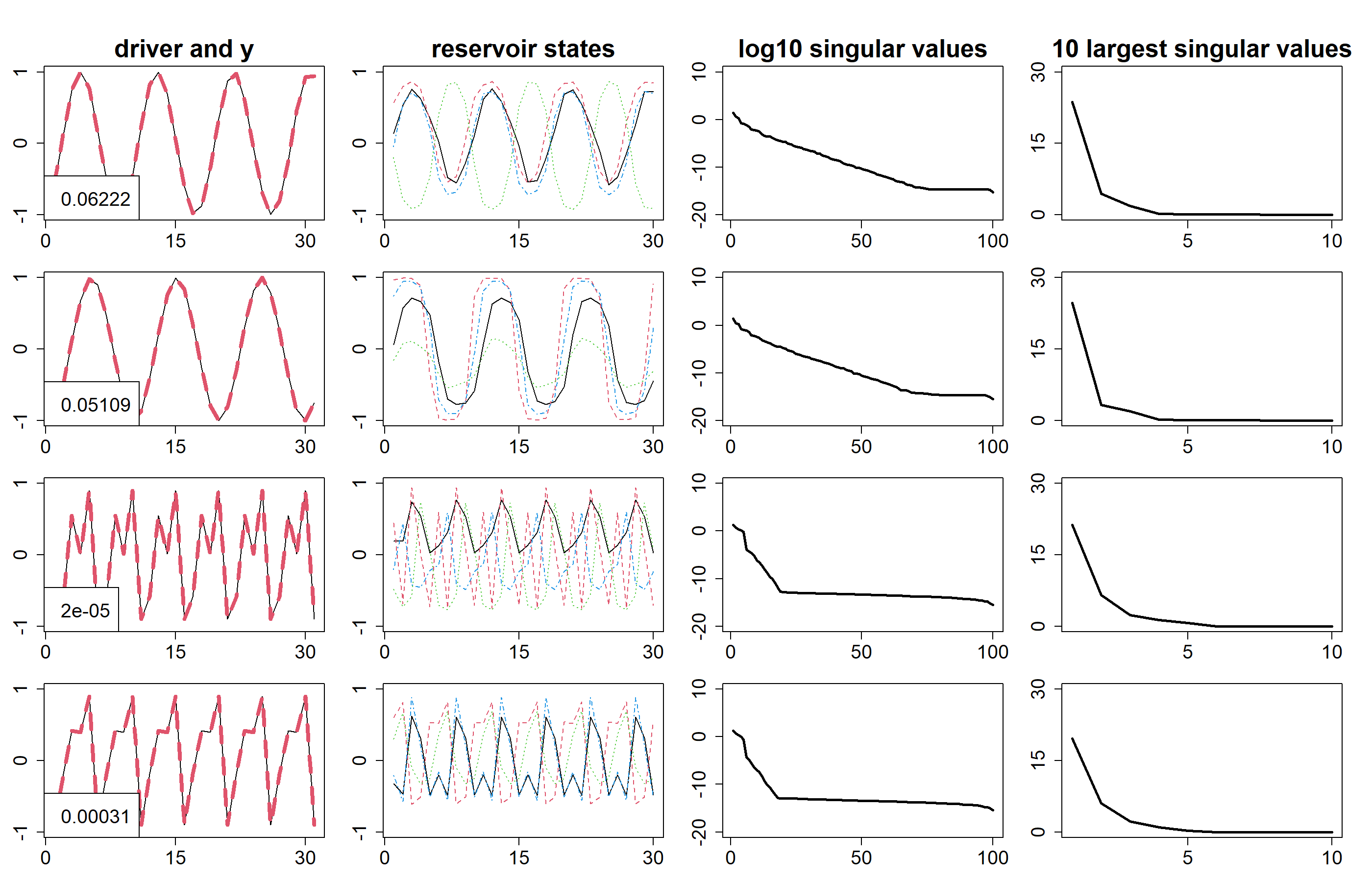}
    \caption{For explanation see Figure \ref{fig:c_pp_results}.}
    \label{fig:dc_pp_results}
\end{figure}

Similar to the conceptors simulations, the reservoir is driven with the four periodic patterns. The diagonal conceptors are computed and the patterns are stored in the reservoir. Afterwards, the reservoir is able to self-generate the patterns with the diagonal conceptors. The self-generated patterns are phase-aligned the same as in the conceptors simulations and compared to the original patterns using NRMSE, see Section \ref{sec:simulation_methods_pattern_comparison}. The results are depicted in Figure \ref{fig:dc_pp_results} and are depicted in a similar way as Figure \ref{fig:c_pp_results}, which allows for direct comparison. Since the meaning of every column has already been discussed in Section \ref{sec:simulations_pp_c}, the interpretation of Figure \ref{fig:dc_pp_results} will not be described again.
\doubleEnter
From the first column of Figure \ref{fig:dc_pp_results} it is evident that diagonal conceptors achieve equally good accuracy as conceptors. Additionally, the second, third, and fourth columns of Figure \ref{fig:dc_pp_results} are approximately identical to the second, third, and fourth columns of Figure \ref{fig:c_pp_results}.
\doubleEnter
Similar to the conceptors simulation, individual parameters were varied to see how it would affect the accuracy. The diagonal conceptors still achieve good accuracy, as defined in Section \ref{sec:simulation_methods_pattern_comparison}, when the parameters are individually modified in the following ranges:
\begin{itemize}
    \item $W^{in}$ scaling: $[0.8,1.05]$ and $[1.15,1.4]$,
    \item $W^{*}$ scaling: $[0.9,1]$,
    \item $\mathbf{b}$ scaling: $[0.15,0.35]$,
    \item $\alpha_{l}$: $[0.85,1]$,
    \item $\varrho^{W^{out}}$: $[0,10000]$,
    \item $\varrho^{W}$: $(0,0.01]$,
    \item $\alpha^{j}$: $[6,9]$, for all $j$.
\end{itemize}
For $\varrho^{W}$ it should be noted that $0$ is not included in the interval due to the singularity in computing the inverse matrix in ridge regression. In addition, it is remarkable that the input weights scaling is split up into two intervals. For values $1.05-1.15$, the simulation yielded diagonal conceptors that were unstable.
\doubleEnter
Again, the above intervals only indicate that the simulation still yields good results if one parameter at a time is modified. Other parameter combinations that yield good accuracy are possible, e.g., $W^{in}$ scaling $=0.6$, $W^{*}$ scaling $=1$, $\mathbf{b}$ scaling $=0.2$, $\alpha_{l}=1$, $\varrho^{W^{out}}=0$, $\varrho^{W}=0.000001$, $\alpha^{j}=25$ for all $j$.

\subsection{Chaotic Attractors}
\label{sec:simulations_chaotic_attractors}
\begin{figure}[ht]
    \centering
    \includegraphics[width=0.8\textwidth, height=12cm]{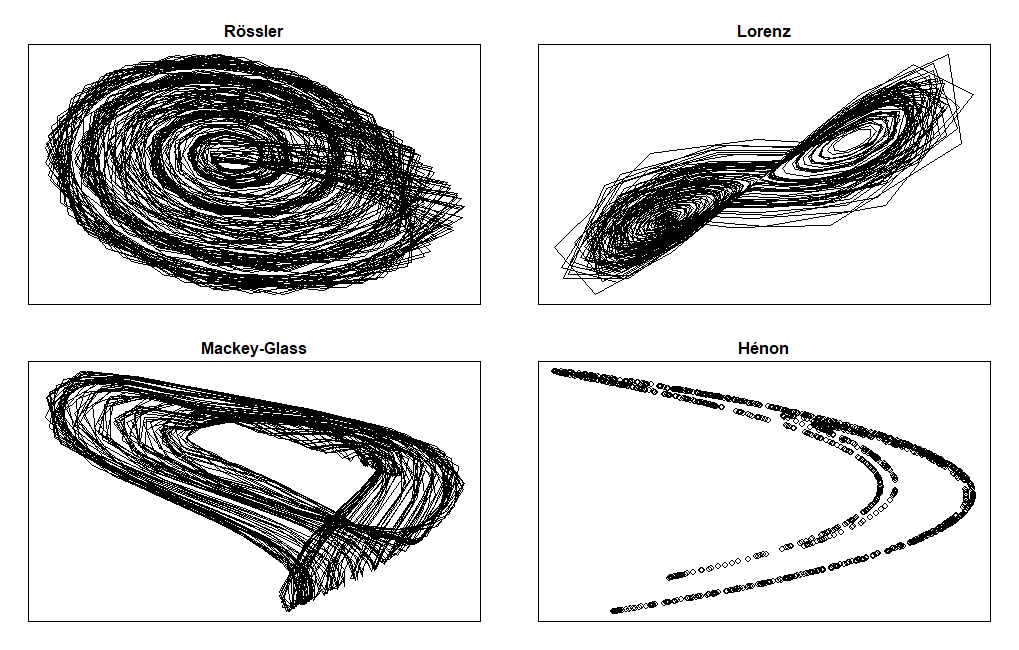}
    \caption{Four chaotic patterns, sampled from the Rössler (top left), Lorenz (top right), Mackey-Glass (bottom left) and Hénon (bottom right) attractors and plotted with time-delay embedding of $1500$ time steps on the interval $[0,1]^{2}$.}
    \label{fig:ca}
\end{figure}

This example comprises four chaotic patterns sampled from the well-known Rössler, Lorenz, Mackey-Glass and Hénon attractors. The patterns are chaotic, hence fragile, and it will require careful fine-tuning of the dynamics of the reservoir in combination with the aperture of the conceptors and diagonal conceptors. For the defining equations of these well-known chaotic attractors, the reader is referred to Section 4.2 of the conceptors report \cite{MonsterReport}. The four chaotic patterns are shown in Figure \ref{fig:ca}.
\doubleEnter
Let the set of patterns be denoted by $\mathbf{P}=\{\mathbf{p^{1}},\mathbf{p^{2}},\mathbf{p^{3}},\mathbf{p^{4}}\}$, where $\mathbf{p}^{j}(n)\in\mathbb{R}^{2}$, for all $j$. Patterns $1$, $2$, $3$, and $4$ were sampled from the \textit{Rössler}, \textit{Lorenz}, \textit{Mackey-Glass}, and \textit{Hénon} attractors, respectively, and projected to the $2$-dimensional plane via the same projections as in the conceptors report \cite{MonsterReport}. Each pattern has length $L=1500$. The number of neurons in the reservoir was set to $N=500$. The input weights $W^{in}$ and the reservoir weights $W^{*}$ were both scaled by $1.5$, the bias $\mathbf{b}$ was scaled by $1$. The leaking rate was set to $\alpha_{l}=1$. The length of the washout period was set to $n_{washout}=100$, and for diagonal conceptors the length of stage 1 was set to $n_{stage1}=400$, yielding a stage 2 length of $n_{stage2}=L-n_{washout}-n_{stage1}=1000$ for diagonal conceptors and $n_{learn}=L-n_{washout}=1400$ for conceptors. For diagonal conceptors, the apertures were set to $\alpha^{1}=10$, $\alpha^{2}=6$, $\alpha^{3}=9$, $\alpha^{4}=5$, where $\alpha^{j}$ corresponds to pattern $j$. For conceptors, the apertures were set to
$\alpha^{1}=140$, $\alpha^{2}=50$, $\alpha^{3}=140$ and $\alpha^{4}=20$. The regularization constant for computing the output weights was set to $\varrho^{W^{out}}=0$ for both conceptors and diagonal conceptors and the regularization constant for the recomputing the reservoir weights was set to $\varrho^{W}=0.1$ for conceptors and $\varrho^{W}=0.01$ for diagonal conceptors.
\doubleEnter
The chaotic behavior of the patterns requires a carefully chosen aperture. Therefore, the goal of this example is to demonstrate how the role of the aperture for diagonal conceptors differs from the role of the aperture for conceptors. The control that the aperture offers for conceptors is slightly lost for diagonal conceptors. The reason is as follows. On the one hand, in the training scheme of conceptors, the conceptor matrices are computed independently of $W$ and $W^{out}$. Therefore, after the reservoir is loaded, an adjustment in the aperture will not affect $W$ nor $W^{out}$. Adjusting the aperture will only affect the conceptor matrices. On the other hand, in the training scheme of diagonal conceptors, the diagonal conceptor matrices are computed and inserted in the update equations \textit{before} the patterns are stored. However, since an adjustment in the aperture changes the diagonal conceptor matrices, that adjustment in the aperture also affects the matrices $W$ and $W^{out}$ and thus the resulting dynamics of the reservoir. Consequently, changing the aperture associated with pattern $j$ may affect the self-generated pattern associated with diagonal conceptor $i\neq j$. As a result, for diagonal conceptors there is a delicate interplay between the aperture and the regularization constants for  $W$ and $W^{out}$.

\subsubsection{Conceptors}
\begin{figure}[ht]
    \centering
    \includegraphics[width=0.9\textwidth]{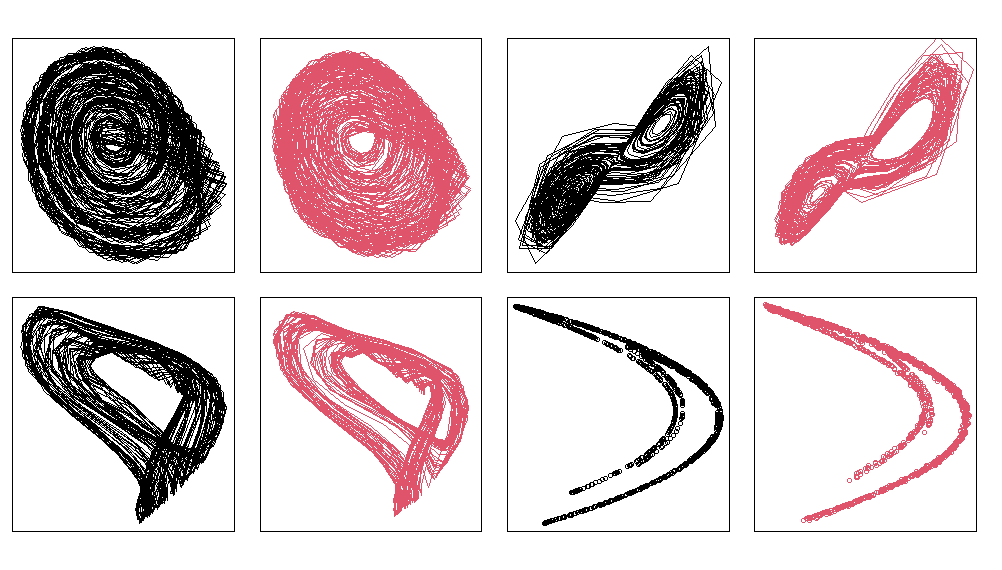}
    \caption{Conceptor-generated patterns (red) plotted next to their target patterns (black) in a time-delay embedding of $1500$ time steps on the interval $[0,1]^{2}$.}
    \label{fig:c_ca_results}
\end{figure}

The conceptors were trained according to the procedure described in Section \ref{sec:training_conceptors}. The resulting conceptor matrices were then used to self-generate the stored patterns. The self-generated patterns were compared to the original patterns by visual inspection, because they are not expected to align perfectly with their target patterns. The target patterns are chaotic in nature, so if the self-generated patterns would be able to predict the chaotic patterns it would defy the unpredictability of the chaotic patterns. There are methods for comparing two chaotic patterns, for example by computing and comparing the Lyapunov spectrums, however, these were not employed in this thesis. Instead, the characteristics of the self-generated patterns and the target patterns were visually inspected to assess the accuracy of the match. In Figure \ref{fig:c_ca_results} it can be seen that the characteristics are indeed captured. The black patterns are the target patterns and the red patterns are the conceptor-generated patterns. Furthermore, they are stable, meaning that after perturbation, the dynamics of the reservoir will return to the dynamics associated with the chaotic attractor.
\doubleEnter
In the periodic patterns simulation, the individual parameters were slightly varied to find in which ranges the self-generated patterns would still be satisfactory. However, in the case of chaotic attractors, this is a bit more complicated. The reason is that the aperture must be handled with care, so changing a parameter such as the bias scaling leads to a necessary adjustment in the apertures. For example, adjusting the input weights scaling from $1.5$ to $1.4$ yields unstable self-generated patterns, but if the apertures of patterns $3$ and $4$ are then also adjusted to $\alpha^{2}=35$ and $\alpha^{3}=65$, the self-generated patterns are stable again.

\subsubsection{Diagonal Conceptors}
\begin{figure}[ht]
    \centering
    \includegraphics[width=0.9\textwidth]{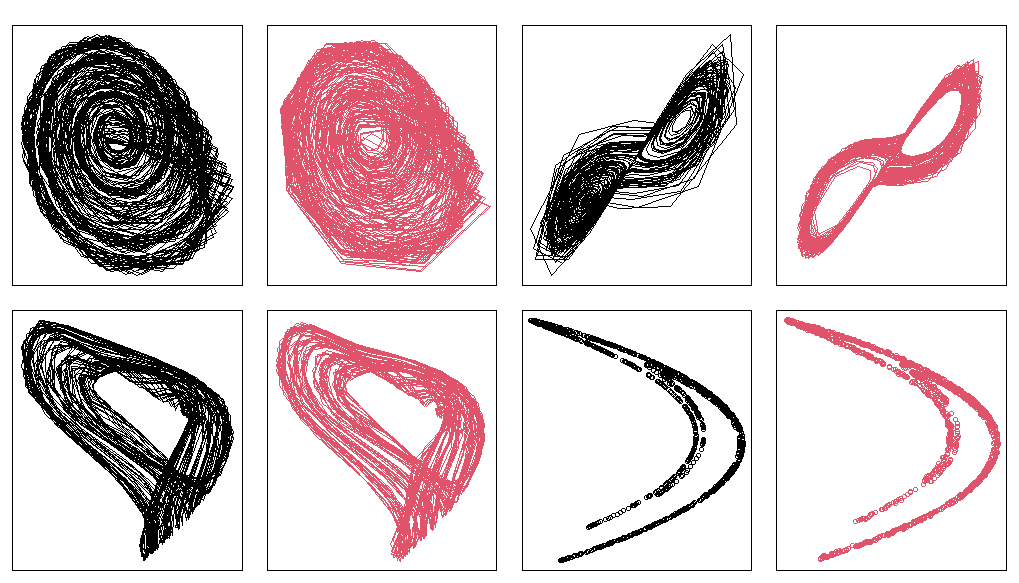}
    \caption{Diagonal conceptor-generated patterns (red) plotted next to their target patterns (black) for a period of $L=1500$ time steps.}
    \label{fig:dc_ca_results}
\end{figure}

The diagonal conceptors were trained according to the scheme described in Section \ref{sec:training_diagonal_conceptors}. The resulting diagonal conceptors were then used to self-generate the patterns. Similar to conceptors, the self-generated patterns were compared to the target patterns by visual inspection, and both the target patterns and the self-generated patterns are shown in Figure \ref{fig:dc_ca_results}. The diagonal conceptor-generated patterns resemble the target patterns, just as well as in the conceptors simulation. Furthermore, they also yield stable reservoir dynamics like conceptors.
\doubleEnter
Similar to the parameters in the conceptors simulation, the parameters in the diagonal conceptors simulation could not simply be adjusted individually. Moreover, the aperture adjustment for the chaotic patterns was more sensitive than for the periodic patterns, i.e., a slight modification could yield unstable results. The main reason is that adjusting the aperture for $j$ may affect the outcome for pattern $i\neq j$, as mentioned in the introduction of this example. For example, adjusting the aperture of pattern $1$ from $\alpha^{1}=10$ to $\alpha^{1}=11$ resulted in instability for pattern $2$, whereas pattern $1$, $3$ and $4$ remained stable. For this reason, it was slightly more difficult to find stable solutions. Once a stable solution has been found for some of the patterns, one cannot simply adjust the aperture of the unstable patterns, because it could disrupt the stability of the stable patterns. In contrast to conceptors, one must start by adjusting the apertures to overall acceptable solutions and then gradually fine-tune until satisfaction. Unfortunately, in the current training procedure, the diagonal conceptors cannot be adjusted one by one. Furthermore, the stability of the solutions is not only dependent on the aperture, but also on the regularization constants $\varrho^{W}$ and $\varrho^{W^{out}}$. There is a subtle balance between $\varrho^{W}$ and the aperture. Intuitively, if the diagonal conceptor-generated patterns seem to be chaotic, either the aperture must be decreased or the regularization constant must be increased. 
\doubleEnter
However, regardless of the increased difficulty of finding the appropriate parameters, there were still many different parameter configurations that yielded stable solutions. Similar to the periodic patterns simulations, some of the parameters could be individually varied while maintaining good accuracy, as described in Section \ref{sec:simulation_methods_pattern_comparison}, in the following ranges:
\begin{itemize}
    \item $W^{in}$ scaling: $[1.5,1.7]$,
    \item $W^{*}$ scaling: $[1.4,1.5]$,
    \item $\mathbf{b}$ scaling: $[0.8,1]$,
    \item $\alpha_{l}$: $[0.9,1]$,
    \item $\varrho^{W^{out}}$: $[0,1]$.
\end{itemize}
First, notice that the ranges are smaller compared to the periodic patterns simulations, which made it initially more difficult to find parameters that yield stable solutions. Second, note that the ranges for $\varrho^{W}$ and $\alpha^{j}$ are omitted. This is due to the fact that even a slight adjustment in either $\varrho^{W}$ or $\alpha^{j}$ would cause the self-generated patterns to be unstable. Lastly, similar to the periodic patterns simulations, other combinations of parameters also yielded good solutions, e.g.,  $W^{in}$ scaling $=1.1$, $W^{*}$ scaling $=1.2$, $\mathbf{b}$ scaling $=0.8$, $\alpha_{l}=1$, $\varrho^{W^{out}}=0.001$, $\varrho^{W}=0.01$, $\alpha^{1}=30$, $\alpha^{2}=30$, $\alpha^{3}=30$, $\alpha^{4}=20$.

\subsection{Human Motion}
\label{sec:simulations_human_motion}
\begin{figure}[ht]
    \centering
    \includegraphics[width=\textwidth]{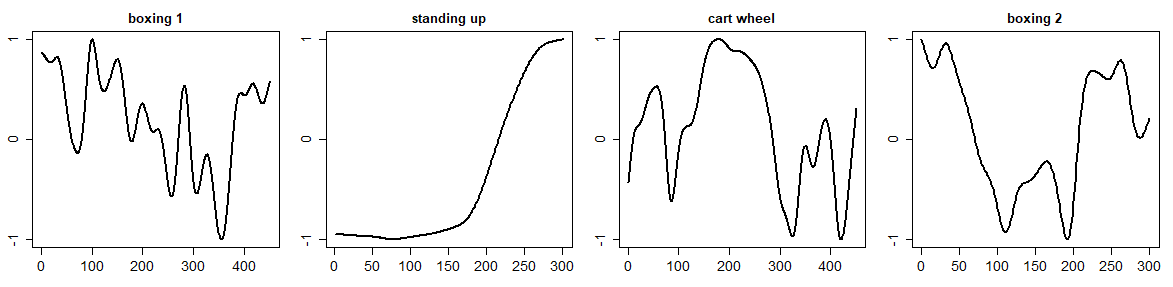}
    \caption{The first of $61$ dimensions is plotted for four different human motion patterns: boxing 1, standing up, cartwheel, and boxing 2.}
    \label{fig:hm}
\end{figure}

The patterns used in the simulations in this section are taken from the supplementary code of the paper \textit{Using Conceptors to Manage Long-Term Memory For Temporal Patterns} by H. Jaeger\cite{ConceptorsHumanMotion}. In this paper, it is shown how conceptors can be used to manage long-term memory for temporal patterns by the use of a few examples. The human motion example is a real-world example demonstrating that conceptors are capable of storing a larger number of patterns of different lengths and timescales. Therefore, this example is ideal for demonstrating that diagonal conceptors are also capable of that. Even though the patterns in this example were taken from the paper, it must be noted that the simulations were also performed with a different set of human motions, which yielded similar results. These simulations can be found at the aforementioned GitHub repository \cite{GithubRepoDiagonalConceptors}.
\doubleEnter
The patterns are derived from real human motion, recorded by a motion capture lab consisting of $12$ infrared camera capable of recording $120$ Hz with images of $4$ megapixel resolution. The data is taken from the open source Carnegie Mellon University Motion Capture Database (MoCap) \cite{MoCap}. A test subject wears a black jumpsuit and has $41$ markers taped to the suit. The cameras pick up the markers and the different images are then used to triangulate the positions of the markers to output 3D data.
\doubleEnter
The data can be used in two ways. Either the data is given in the form of a .c3d file, which contains the marker positions, or the data is given by a pair of .asf and .amc files, where the .asf file describes the skeleton and its joints and the .amc file describes the movement data. Since the data is in a specific format, it is not so straightforward to extract the human motion from the data. Fortunately, code has been written in different programming languages to accommodate this. The patterns in the supplementary code of \cite{ConceptorsHumanMotion} are already processed for the most part, but in the case of data taken directly from the MoCap database, an R package called \textit{mocap} (\url{https://github.com/gsimchoni/mocap}) was used for parsing the .asf and .amc files. For a more detailed explanation about the structure of the data, the reader is referred to the MoCap website (\url{http://mocap.cs.cmu.edu/}). Since the data processing is rather involved, it will not be discussed here, but in the case of the patterns that are used in this section, the reader can read more about it in the appendix of \cite{ConceptorsHumanMotion}.
\doubleEnter
Let $\mathbf{P}=\{\mathbf{p}^{1}, \mathbf{p}^{2},...,\mathbf{p}^{15}\}$ be the set of patterns where pattern $j$ at time step $n$ is given by $\mathbf{p}^{j}(n)\in\mathbb{R}^{61}$, for all $j$. Patterns $1$ until $15$ represent the following human motions, respectively: three different types of boxing, doing a cartwheel, crawling, striding, getting down on the knees, getting seated, jogging, sitting, walking slowly, standing up, standing up from a stool, walking, waltzing. The patterns are diverse in their length $L$, which ranges between roughly $200$ and $900$ time steps. Some are transient (sitting down), some are periodic (walking, running), and some are irregular stochastic (boxing). The columns of each pattern are scaled such that they lie in the range $[-1,1]$. In addition, the patterns were smoothed to remove some of the rough edges of the data. The first dimension of four patterns is shown in Figure \ref{fig:hm}. 
\doubleEnter
The goal of this example is to demonstrate the ability of diagonal conceptors for a larger, more diverse set of patterns. The fact that this is a real-world example makes it a fitting challenge for diagonal conceptors.

\subsubsection{Conceptors}
\begin{figure}[ht]
    \centering
    \includegraphics[width=\textwidth]{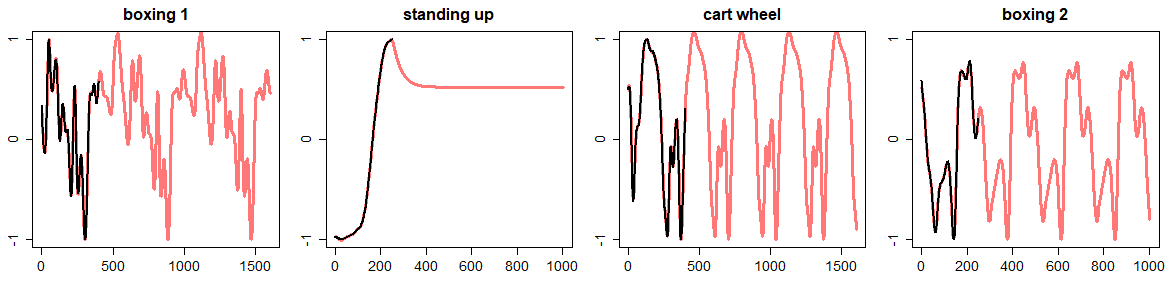}
    \caption{The first of $61$ dimensions is plotted for the conceptor-generated patterns (red) as well as the target pattern (black), where the conceptor-generated pattern is run for a period that is $4$ times longer than its target pattern. The human motions associated with the four different patterns are: boxing 1, standing up, cartwheel, and boxing 2.}
    \label{fig:c_hm_results}
\end{figure}

The number of neurons in the reservoir was set to $N=600$. The input weights $W^{in}$ were scaled by $0.2$, the bias $\mathbf{b}$ was scaled by $0.8$ and the reservoir weights $W^{*}$ were scaled by $1$. The leaking rate was set to $\alpha_{l}=0.3$. The length of the washout period was set to $n_{washout}=50$, yielding a learning period of length $n_{learn}=L-50$, which was different for each patterns. The apertures were set to $\alpha^{1}=25$, $\alpha^{2}=55$, $\alpha^{3}=17$, $\alpha^{4}=49$, $\alpha^{5}=89$, $\alpha^{6}=49$, $\alpha^{7}=9$, $\alpha^{8}=65$, $\alpha^{9}=49$, $\alpha^{10}=9$, $\alpha^{11}=25$, $\alpha^{12}=25$, $\alpha^{13}=17$, $\alpha^{14}=9$, and $\alpha^{15}=7$, where each $\alpha^{j}$ corresponds to the aperture for pattern $j$. The apertures were found by grid search where the grid was set from $1$ to $100$ with intervals of $9$, where the aperture for pattern $15$ was adjusted after the grid search. The regularization constants were set to $\varrho^{W^{out}}=0$ and $\varrho^{W}=0.001$.
\doubleEnter
The patterns were stored in the reservoir and the conceptor matrices were computed. With the conceptor matrices, the reservoir was able to self-generate the patterns from a given starting reservoir state. This is different from the periodic patterns and chaotic attractors simulations, where the reservoir could start from any reservoir state. The conceptor-generated patterns would sometimes show unpredictable behavior if the reservoir would start from a random initial reservoir state, which is expected from, e.g., the transient patterns, as they must start from the correct starting state. For example, going from a sitting position to a standing position requires that the starting point is the sitting position. Therefore, the starting state of the learning period was saved and used as the starting state of the reservoir for the self-generation of the patterns. The reservoir was run for a period $4$ times longer than the target pattern and the first dimensions of the outputted patterns are shown in Figure \ref{fig:c_hm_results}.
\doubleEnter
For each target pattern and conceptor-generated pattern, the NRMSE is computed along each dimension for the overlapping period, i.e., the first $n_{learn}$ time steps. This yields a vector of $61$ NRMSEs per pattern, of which the minimum, maximum, mean and standard deviation are shown in Table \ref{tab:c_hm}. The table shows that the mean NRMSE of each pattern, except for pattern $15$, is below $0.1$, which is desired. Pattern $15$ corresponds to the waltz human motion, which is the longest pattern out of the $15$ patterns with a length of $900$ time steps. As mentioned in Section \ref{sec:initializing_the_parameters}, the rule of thumb for initializing $N$ is to set is equal to the length of the longest pattern. However, in this case, $N=600$ and the longest pattern has length $900$, which is assumed to be the reason that the NRMSE for pattern $15$ is slightly larger. Furthermore, the small standard deviation row indicates that the NRMSE stay roughly around the mean NRMSE across all $61$ dimensions, i.e., the mean is representative of the overall NRMSE.
\doubleEnter
\begin{table}[ht]
\centering
\begin{tabular}{|l|l|l|l|l|l|l|l|l|l|}
\hline
Pattern & 1 & 2 & 3 & 4 & 5 & 6 & 7 & 8 & 9\\ \hline
\hline
Min & 0.058 & 0.061 & 0.030 & 0.021 & 0.041 & 0.039 & 0.012 & 0.009 & 0.070\\ \hline
Max & 0.123 & 0.183 & 0.099 & 0.074 & 0.141 & 0.108 & 0.077 & 0.096 & 0.202\\ \hline
\textbf{Mean} & \textbf{0.076} & \textbf{0.092} & \textbf{0.047} & \textbf{0.036} & \textbf{0.057} & \textbf{0.058} & \textbf{0.030} & \textbf{0.023} & \textbf{0.083}\\ \hline
Std & 0.017 & 0.025 & 0.016 & 0.012 & 0.023 & 0.018 & 0.017 & 0.016 & 0.030\\ \hline
\end{tabular}
\bigskip
\vspace{1mm}
\begin{tabular}{|l|l|l|l|l|l|}
\hline
10 & 11 & 12 & 13 & 14 & 15 \\ \hline
\hline
0.031 & 0.041 & 0.023 & 0.008 & 0.058 & 0.123\\ \hline
0.237 & 0.107 & 0.117 & 0.161 & 0.289 & 0.466\\ \hline
\textbf{0.060} & \textbf{0.056} & \textbf{0.039} & \textbf{0.025} & \textbf{0.084} & \textbf{0.234}\\ \hline
0.039 & 0.017 & 0.022 & 0.023 & 0.042 & 0.084\\ \hline
\end{tabular}
\caption{The NRMSE of the target patterns and the conceptor-generated patterns. The NRMSE is computed per dimension, which yields a vector of $61$ NRMSEs. The Mean row shows the average NRMSE. The Min row shows the smallest NRMSE. The Max row shows the largest NRMSE. The Std row shows the standard deviation of the NRMSEs vector.}
\label{tab:c_hm}
\end{table}
The admissible ranges of the parameters were less broad than the admissible ranges of the parameters in previous simulations. For example, changing the input weights scaling from $0.2$ to $0.3$ disrupts the conceptors associated with pattern $2$ and $15$. However, the conceptors associated with the disrupted patterns can often be recomputed with different apertures, without changing the conceptors of the other patterns. This way, if the balance is disrupted, it can be restored by adjusting the apertures accordingly. 
\doubleEnter
Furthermore, the conceptors are moderately robust to unknown reservoir states, which is why it has no trouble repeating a periodic pattern, as is seen Figure \ref{fig:c_hm_results}. This is because the starting reservoir state for a periodic human motion will be roughly the same as the ending reservoir state, e.g., a cartwheel starts from a standing position and ends in an ending position. However, in the case of transient patterns, the long-term behavior is undefined, which is why the self-generated pattern will show unpredictable behavior after the pattern has been regenerated. The long-term behavior of the self-generated patterns for both the periodic and transient patterns can be seen in Figure \ref{fig:c_hm_results}. Notice how the periodic human motion repeats after the target pattern has been regenerated, but the transient pattern stagnates.

\subsubsection{Diagonal Conceptors}
\begin{figure}[ht]
    \centering
    \includegraphics[width=\textwidth]{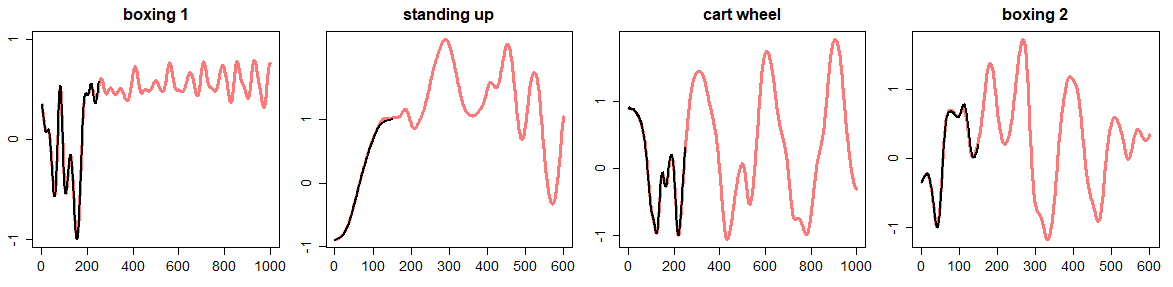}
    \caption{The first of $61$ dimensions is plotted for the diagonal conceptor-generated output (red) as well as the target pattern (black), where the conceptor-generated patterns are run for a period that is $4$ times longer than the target pattern. The human motion associated with the four different patterns are: boxing 1, standing up, cartwheel, and boxing 2.}
    \label{fig:dc_hm_results}
\end{figure}

The number of neurons in the reservoir was set to $N=1000$. The input weights $W^{in}$ were scaled by $0.1$, the bias $\mathbf{b}$ was scaled by $0.1$ and the reservoir weights $W^{*}$ were scaled by $1$. The leaking rate was set to $\alpha_{l}=0.15$. The length of the washout period was set to $n_{washout}=50$ and the length of stage 1 $n_{stage1}$ was set to the ceiling of one-third of the total length of the patterns, except for pattern $15$, where a slightly larger period of $n_{stage1}=400$ was required. This resulted in a stage 2 length $n_{stage2}=L-n_{stage1}-n_{washout}$, which was different for every pattern. The apertures were set to $\alpha^{1}=8$, $\alpha^{2}=8$, $\alpha^{3}=20$, $\alpha^{4}=8$, $\alpha^{5}=35$, $\alpha^{6}=35$, $\alpha^{7}=35$, $\alpha^{8}=8$, $\alpha^{9}=35$, $\alpha^{10}=35$, $\alpha^{11}=35$, $\alpha^{12}=35$, $\alpha^{13}=30$, $\alpha^{14}=35$, $\alpha^{15}=8$, where each $\alpha^{j}$ corresponds to the aperture for pattern $j$. The regularization constants were set to $\varrho^{W^{out}}=0.1$ and $\varrho^{W}=0.05$.
\doubleEnter
After the patterns were stored in the reservoir and the diagonal conceptors were computed, the reservoir was run freely with diagonal conceptors in the loop, where, again, the starting reservoir state was given. The diagonal conceptor-generated patterns are run for a period of length $4*n_{stage2}$, similar to the conceptors simulation, and the first dimension of a few of the diagonal conceptor-generated patterns are shown in Figure \ref{fig:dc_hm_results}. The black patterns show the training patterns and the red patterns show the diagonal conceptor-generated patterns.
\doubleEnter
First of all, notice that stage 1 shortens the training patterns. For the periodic patterns and the chaotic attractors, this was not a problem, because the data was synthetically generated and stage 1 could simply be generated to fit the needs of the simulation. However, for these real-world patterns this becomes a bit of an issue, because stage 1 is required to train the diagonal conceptors. Ideally, the characteristics of a pattern are fully captured in stage 1 and then repeated in stage 2. Unfortunately, this is sometimes not the case for real-world examples. One way to overcome this would be to reuse the pattern from stage 1 in stage 2. This is discussed in more detail in Section \ref{sec:stage1}.
\doubleEnter
Second of all, from Figure \ref{fig:dc_hm_results} it seems that, after the target pattern has been matched, the diagonal conceptors-generated patterns are not fully behaving like they are expected to. The diagonal conceptors are trained on $n_{stage2}$ reservoir states, so starting from a given reservoir state the diagonal conceptors can constrain the dynamics reservoir as they were taught for $n_{stage2}$ time steps. However, after those $n_{stage2}$ time steps, the reservoir state is unknown to the diagonal conceptors. At that point, the diagonal conceptors should force the reservoir state back to the reservoir state it knows. In the conceptors simulation it was seen that the conceptors were more robust to unknown reservoir states. The diagonal conceptors are more vulnerable to unknown reservoir states than conceptors, as is seen by the fact that the diagonal conceptors-generated patterns show unpredictable behavior after the reservoir has autonomously run for $n_{stage2}$ time steps. In other words, diagonal conceptors are less powerful than conceptors to guide the reservoir back from unknown states to states that it knows.
\doubleEnter
Similar to the conceptors simulation, the patterns were compared for the first $n_{stage2}$ steps, corresponding to the black line in Figure \ref{fig:dc_hm_results}, and the NRMSEs are shown in Table \ref{tab:dc_hm}. The results are similar to the results for conceptors.
\begin{table}[ht]
\centering
\begin{tabular}{|l|l|l|l|l|l|l|l|l|l|}
\hline
Pattern & 1 & 2 & 3 & 4 & 5 & 6 & 7 & 8 & 9\\ \hline
\hline
Min & 0.013 & 0.009 & 0.020 & 0.005 & 0.005 & 0.023 & 0.009 & 0.015 & 0.004\\ \hline
Max & 0.058 & 0.064 & 0.254 & 0.037 & 0.136 & 0.473 & 0.364 & 0.534 & 0.051\\ \hline
\textbf{Mean} & \textbf{0.025} & \textbf{0.020} & \textbf{0.060} & \textbf{0.011} & \textbf{0.013} & \textbf{0.090} & \textbf{0.026} & \textbf{0.061} & \textbf{0.010}\\ \hline
Std & 0.011 & 0.009 & 0.044 & 0.005 & 0.022 & 0.067 & 0.052 & 0.127 & 0.011\\ \hline
\end{tabular}
\bigskip
\vspace{1mm}
\begin{tabular}{|l|l|l|l|l|l|}
\hline
10 & 11 & 12 & 13 & 14 & 15 \\ \hline
\hline
0.010 & 0.011 & 0.008 & 0.019 & 0.045 & 0.017\\ \hline
0.339 & 0.146 & 0.342 & 1.308 & 0.175 & 0.107\\ \hline
\textbf{0.036} & \textbf{0.043} & \textbf{0.012} & \textbf{0.091} & \textbf{0.086} & \textbf{0.054}\\ \hline
0.075 & 0.032 & 0.061 & 0.184 & 0.033 & 0.021\\ \hline
\end{tabular}
\caption{The NRMSE of the target patterns and the diagonal conceptor-generated patterns. The NRMSE is computed per dimension, which yields a vector of $61$ NRMSEs. The Mean row shows the average NRMSE. The Min row shows the smallest NRMSE. The Max row shows the largest NRMSE. The Std row shows the standard deviation of the NRMSEs vector.}
\label{tab:dc_hm}
\end{table}
The mean of the NRMSE of each pattern is smaller than $0.1$, which is desired, and the standard deviation is also smaller than $0.1$, indicating that the mean is representative of the performance over most dimensions. Nonetheless, compared to Table \ref{tab:c_hm}, there are a few more outliers, as can be seen in the Max row. Especially pattern $8$, which has one dimension in which the NRMSE is incredibly large. However, upon closer inspection, this concerns an individual outlier, which only diverges slightly towards the end. Overall, the diagonal conceptors achieve good accuracy.
\doubleEnter
The admissible range of the parameters is similar to, but slightly more fragile than, the admissible range of the parameters of the conceptors simulation. For example, changing the input weights scaling from $0.1$ to $0.2$ disrupts the diagonal conceptors associated with patterns $2$ and $15$, similar to conceptors. Again, this disruption can then be countered by adjusting the apertures $\alpha^{2}$ and $\alpha^{15}$. However, unlike in the case of conceptors, the aperture adjustment of the diagonal conceptors is a bit more delicate, as described in the chaotic attractors simulation. Therefore, adjusting apertures $\alpha^{2}$ and $\alpha^{15}$ may lead to disruption in other diagonal conceptors as well. It is for this reason that it can be difficult to find the right parameters for diagonal conceptors when it involves a larger number of patterns with diverse lengths and higher dimensionality. Regardless, other parameter combinations were found that achieved good accuracy.

\subsection{Morphing}
\label{sec:simulations_morphing}
In this section, the conceptors and diagonal conceptors that were computed in the previous examples were used to morph between different patterns. All examples of morphing in this section are first carried out with conceptors and then with diagonal conceptors to allow for direct comparison.
\doubleEnter
The first example (periodic patterns) is also demonstrated in the conceptors report and it shows three types of morphing: frequency, shape, and heterogeneous \cite{MonsterReport}. The first type morphs two sine waves with different periods, the second type morphs two $5$-periodic patterns and the third type morphs a $5$-periodic pattern into a sine wave. The second example (chaotic attractors) shows a morph between two chaotic patterns, specifically, the Rössler and Mackey-Glass attractors. The third example (human motions) demonstrates a morph between two human motions, boxing and a cartwheel.
\doubleEnter
As discussed in Sections \ref{sec:conceptors_morphing} and \ref{sec:diagonal_conceptors_morphing}, a morph between two or more patterns is governed by the mixture parameter $\mu\in\mathbb{R}$, which is typically between $0$ and $1$. In this example, a morph will be only between two patterns, however this is not necessary in general. Let the patterns be denoted by $j_{1}$ and $j_{2}$ with their associated conceptors $C^{j_{1}}$ and $C^{j_{2}}$, or diagonal conceptors $D^{j_{1}}$ and $D^{j_{2}}$. Pattern $j_{1}$ can be linearly morphed into pattern $j_{2}$ by conceptor matrices $C^{j_{1}}$ and $C^{j_{2}}$ according to 
\begin{equation}
\label{eq:simulations_conceptors_morphing}
\begin{aligned}
    \mathbf{r}(n+1)&=\tanh{(W\mathbf{z}(n)+\mathbf{b})},\\
    \mathbf{z}(n+1)&=\big((1-\mu)C^{j_{1}} + \mu C^{j_{2}}\big)\mathbf{r}(n+1),
\end{aligned}
\end{equation}
or by diagonal conceptor matrices $D^{j_{1}}$ and $D^{j_{2}}$ according to
\begin{equation}
\label{eq:simulations_diagonal_conceptors_morphing}
\begin{aligned}
    \mathbf{r}(n+1)&=\tanh{(W\mathbf{z}(n)+\mathbf{b})},\\
    \mathbf{z}(n+1)&=\big((1-\mu)D^{j_{1}} + \mu D^{j_{2}}\big)\mathbf{r}(n+1),
\end{aligned}
\end{equation}
where the leaking rate is omitted from the equations for notational ease. Note that $0\leq\mu\leq1$ indicates interpolation and $\mu<0$ and $\mu>1$ indicate extrapolation. The simulations in this section start by driving the reservoir with a constant $\mu=\mu_{min}$ for a period of time. Then the reservoir is continued to be driven while $\mu$ is gradually increased for a period of $n_{morph}$ time steps until $\mu=\mu_{max}$.

\subsubsection{Periodic Patterns}
\begin{figure}[p]
\centering
\includegraphics[width=0.95\textwidth]{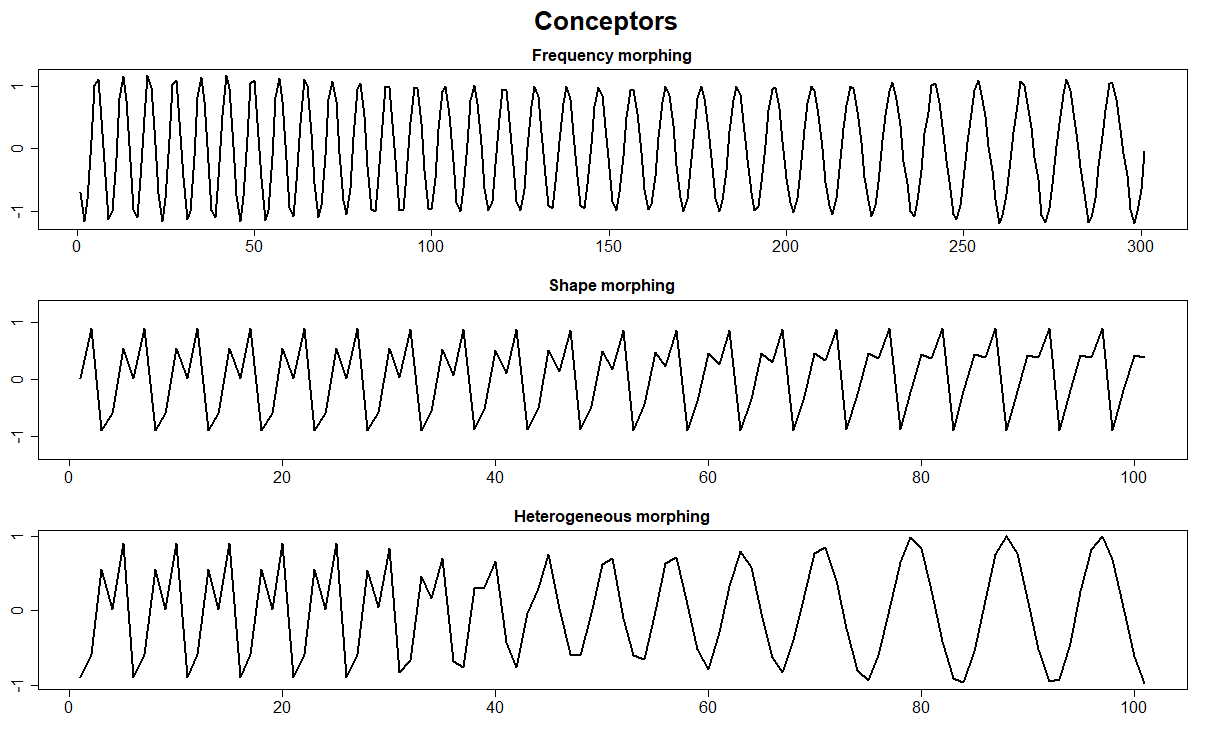}
    \caption{The upper plot shows a morph between two sine waves with periods of approximately $8.83$ and $9.93$, where $-2\leq\mu\leq3$. The middle plot shows a morph between two similar $5$-periodic patterns, where $0\leq\mu\leq1$. The bottom plot shows a morph between a sine wave and a $5$-periodic pattern, where $0\leq\mu\leq1$.}
    \label{fig:c_pp_morphing}
\includegraphics[width=0.95\textwidth]{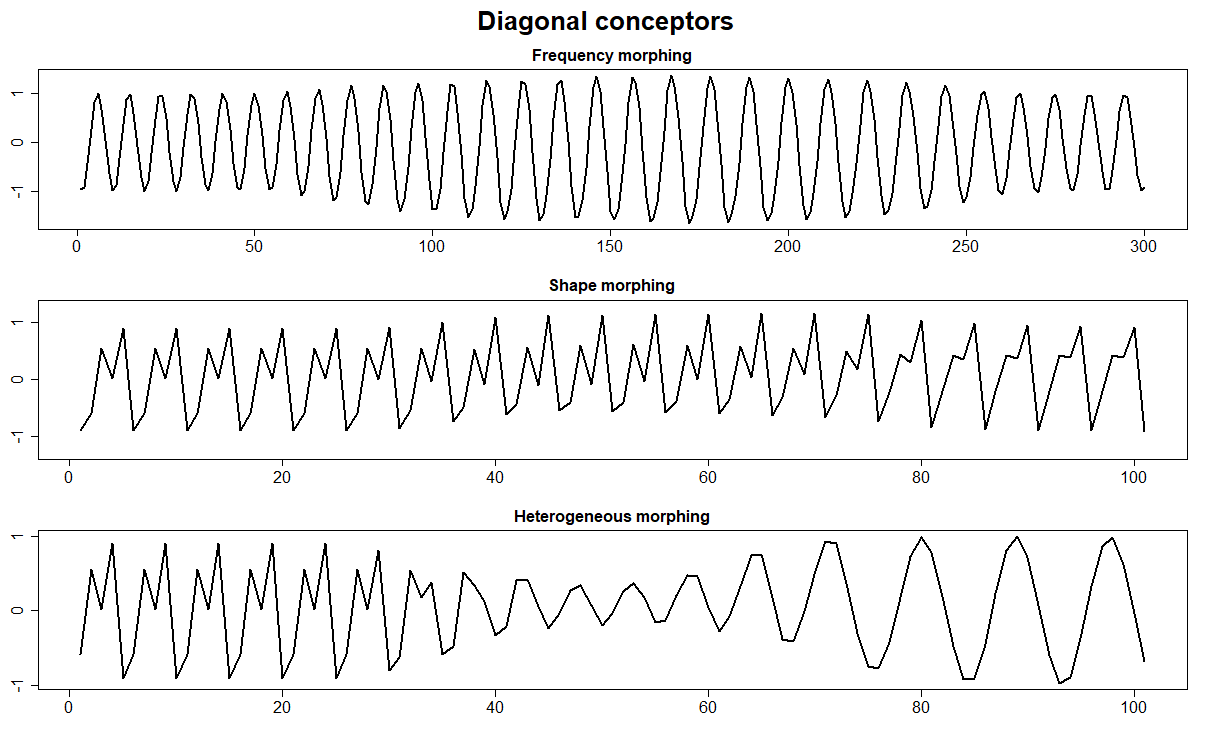}
    \caption{The upper plot shows a morph between two sine waves with periods of approximately $8.83$ and $9.93$, where $0\leq\mu\leq1$. The middle plot shows a morph between two similar $5$-periodic patterns, where $0\leq\mu\leq1$. The bottom plot shows a morph between a sine wave and a $5$-periodic pattern, where $0\leq\mu\leq1$.}
    \label{fig:dc_pp_morphing}
\end{figure}

For morphing conceptors, the boundaries of the mixture parameter were set to $\mu_{min}=-2$ and $\mu_{max}=3$ in the case of for the frequency morphing, and $\mu_{min}=0$ and $\mu_{max}=1$ for the other two types. For morphing diagonal conceptors, the boundaries were set to $\mu_{min}=0$ and $\mu_{max}=1$ for all three types. The reservoir was run according to Equation \ref{eq:simulations_conceptors_morphing} for conceptors and Equation \ref{eq:simulations_diagonal_conceptors_morphing} for diagonal conceptors, where $\mu$ was initialized as $\mu=\mu_{min}$. The first $n_{washout}=100$ steps were discarded. Then, the reservoir was continued to be driven for $50$ steps. Afterwards, $\mu$ was linearly increased until $\mu_{max}$ for $n_{morph}$ time steps. The morphing length was set to $n_{morph}=200$ for the frequency morphing and $n_{morph}=50$ for the others. Finally, the reservoir was run for another $50$ steps with $\mu=\mu_{max}$. The resulting morphed patterns for conceptors and diagonal conceptors are shown in Figure \ref{fig:c_pp_morphing} and \ref{fig:dc_pp_morphing}, respectively.
\doubleEnter
The results for conceptors are similar to the results in the conceptors report, so for more details, the reader is referred to Section 3.7 of the conceptors report \cite{MonsterReport}. In short, in the case of the two sine waves, the conceptors are not only able to interpolate between $\mu=0$ and $\mu=1$, but also extrapolate for values $\mu<0$ and $\mu>1$. Remarkably, the information about the patterns that is extrapolated is the period of the sine waves. Therefore, for $\mu=-2$, the period of the sine wave is approximately $7.34$ and for $\mu=3$, the period is approximately $12.6$, whereas the periods of the original patterns are approximately $8.83$ and $9.83$. This extrapolation feature does not translate to diagonal conceptors. The morphed pattern shows unpredictable behavior for $\mu_{min}<0$ and $\mu_{max}>1$. Furthermore, morphing the two $5$-periodic patterns, as seen in the middle plot of Figures \ref{fig:c_pp_morphing} and \ref{fig:dc_pp_morphing}, shows that diagonal conceptors are just as good as conceptors to morph between similar shapes. Lastly, as seen in the bottom plot of Figures \ref{fig:c_pp_morphing} and \ref{fig:dc_pp_morphing}, diagonal conceptors can smoothly transition between dissimilar shapes.
\doubleEnter
Overall, diagonal conceptors morph between these periodic patterns almost as well as conceptors. However, in contrast to conceptors, they do not facilitate extrapolation.

\subsubsection{Chaotic Attractors}
\begin{figure}[p]
    \centering
    \includegraphics[width=0.95\textwidth]{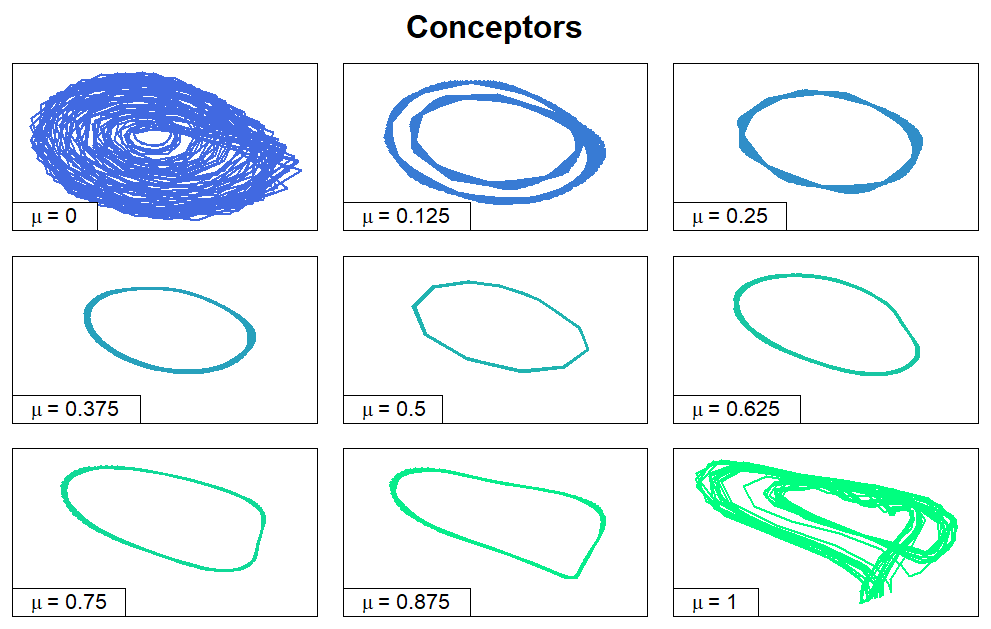}
    \caption{Time-delay embedding of $500$ data points of a morph between the Rössler pattern and the Mackey-Glass pattern, where the mixing parameter $\mu$ is held still at $9$ equidistant values from $\mu=0$ to $\mu=1$. The top left plot shows the Rössler pattern and the bottom right plot shows the Mackey-Glass pattern.}
    \label{fig:c_ca_morphing}
    \includegraphics[width=0.95\textwidth]{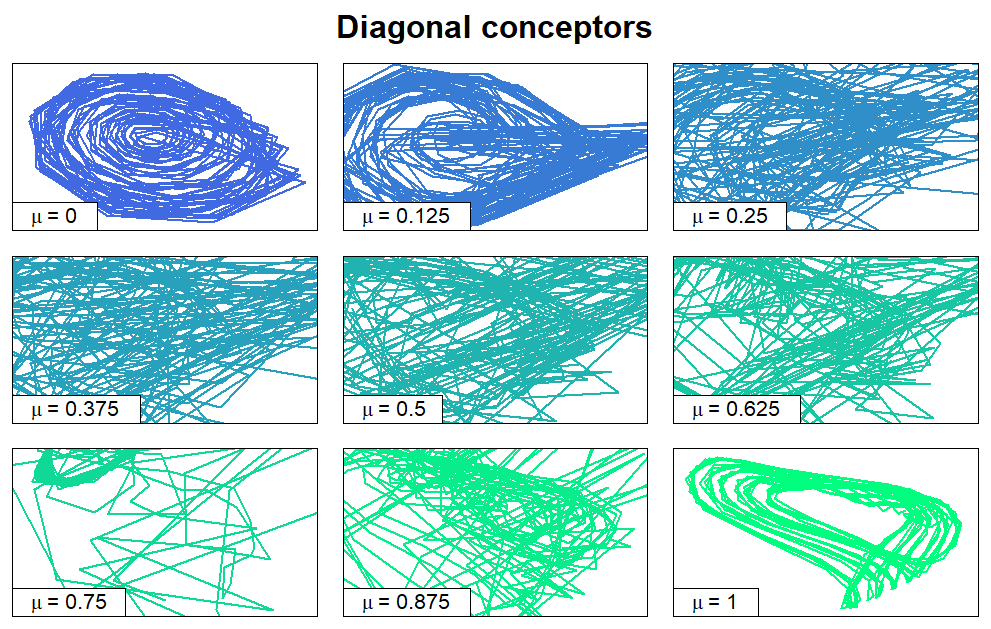}
    \caption{For explanation see Figure \ref{fig:c_ca_morphing}.}
    \label{fig:dc_ca_morphing}
\end{figure}

The reservoir was run according to Equation \ref{eq:simulations_conceptors_morphing} for conceptors and Equation \ref{eq:simulations_diagonal_conceptors_morphing} for diagonal conceptors, where the mixing parameter was initialized as $\mu=0$ and step-wise increased by $0.125$ until $\mu=1$, yielding $9$ values for $\mu$. For every value of $\mu$, the first $n_{washout}=100$ steps were discarded and the reservoir was continued to be run for another $n_{morph}=500$ time steps. For each value of $\mu$, a time-delay embedding of $500$ time steps is shown, resulting in $9$ plots. The resulting plots for conceptors and diagonal conceptors are shown in Figures \ref{fig:c_ca_morphing} and \ref{fig:dc_ca_morphing}, respectively.
\doubleEnter
Both Figures \ref{fig:c_ca_morphing} and \ref{fig:dc_ca_morphing} must be read from the top left to the bottom right plot. The value of $\mu$ corresponding to the plot is shown in the bottom left corner. Notice that $\mu=0$ means that the reservoir is run with only the (diagonal) conceptor corresponding to the Rössler pattern and for $\mu=1$ the reservoir is run with only the (diagonal) conceptor corresponding to the Mackey-Glass pattern. The plots show that interpolation, i.e., $0<\mu<1$, of conceptor matrices yields much more predictable behavior than the interpolation of diagonal conceptor matrices. The diagonal conceptors seem to output unpredictable behavior for $0<\mu<1$. 
\subsubsection{Human Motion}
\begin{figure}[p]
    \centering
    \includegraphics[width=0.95\textwidth]{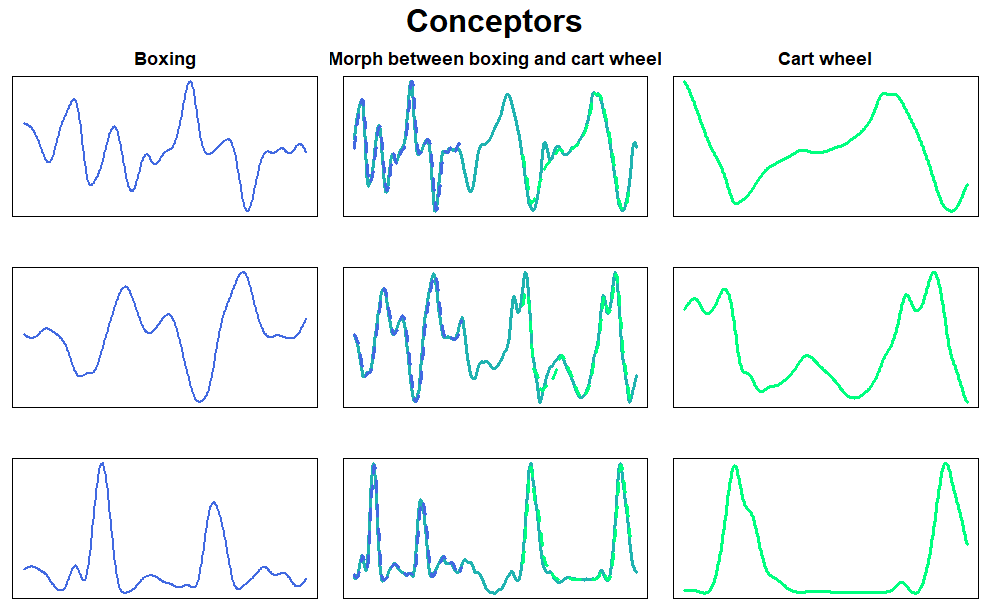}
    \caption{Morphing between two human motions, a boxing motion and a cartwheel motion. Since the patterns are $61$-dimensional, only the first $3$ dimensions are shown, each corresponding to a row in the figure. The left column shows the boxing patterns and the right column shows the cartwheel pattern. The middle column shows how the boxing pattern is linearly morphed into the cartwheel pattern and the dashed lines show the training patterns, which are added to compare the outputs.}
    \label{fig:c_hm_morphing}
    \includegraphics[width=0.95\textwidth]{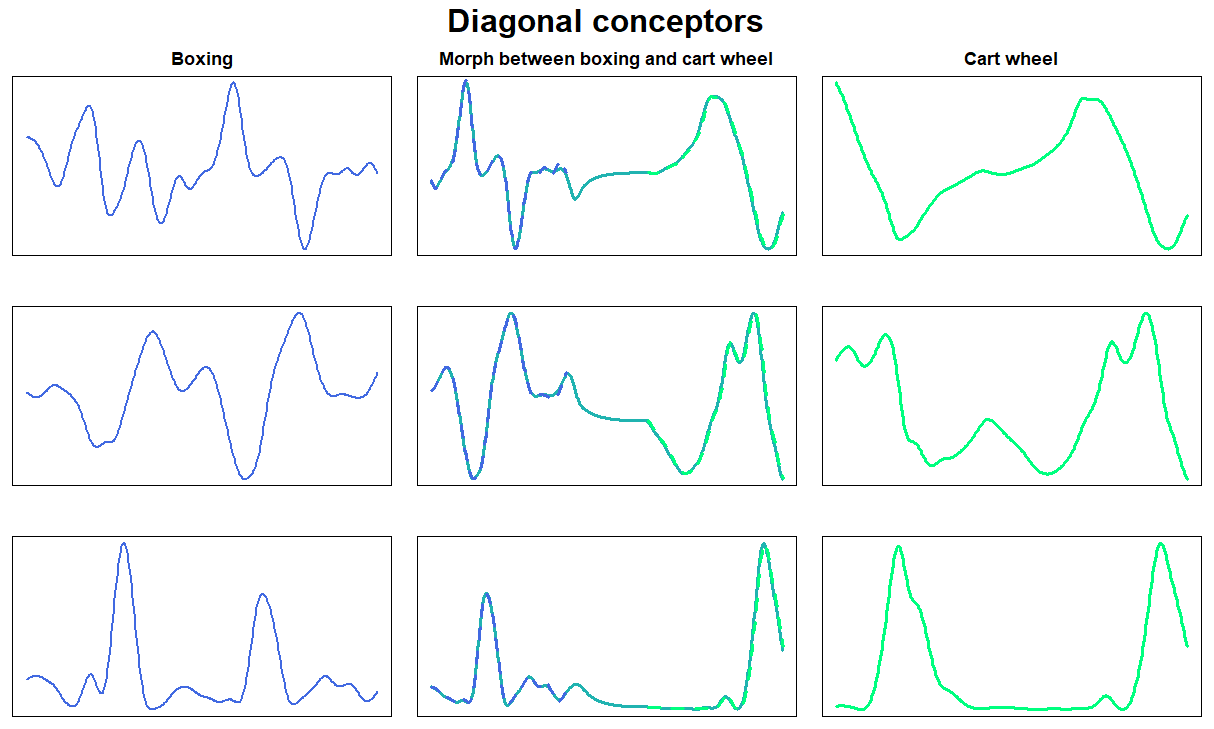}
    \caption{For explanation see Figure \ref{fig:c_hm_morphing}.}
    \label{fig:dc_hm_morphing}
\end{figure}

Again, the reservoir was run according to Equation \ref{eq:simulations_conceptors_morphing} for conceptors and Equation \ref{eq:simulations_diagonal_conceptors_morphing} for diagonal conceptors. The (diagonal) conceptor matrices ($D^{j_{1}}$) $C^{j_{1}}$ and ($D^{j_{2}}$) $C^{j_{2}}$ correspond to a boxing pattern and a cartwheel pattern, respectively. The starting reservoir state was given, so there was no washout period required. The boundaries of the mixture parameter were given by $\mu_{min}=0$ and $\mu_{max}=1$. The reservoir was run with $\mu=0$ for the length of the training pattern $j_{1}$. Then the pattern was morphed for $n_{morph}=150$ time steps. In reality, this would amount to a little more than $1$ second, which is a fair amount of time needed between boxing jabs and doing a cartwheel. After the morphing period, the reservoir was run for another period with the length equal to that of the training pattern $j_{2}$.
\doubleEnter
Remember that the diagonal conceptors were unstable to unknown reservoir states. Therefore, in the case of diagonal conceptors, during the morphing period, the reservoir state was slightly nudged to the correct starting state. This was implemented by adjusting the reservoir state $\mathbf{z}(n)$ as follows during the morphing period: 
\begin{equation}
    \mathbf{z}(n)=(1-\mu)\mathbf{z}(n) + \mu\mathbf{z}^{j_{2}}(0),
\end{equation}
where $\mathbf{z}^{j_{2}}(0)$ is the starting state of pattern $j_{2}$, and $\mu$ is the same mixing parameter as for morphing. This nudging procedure was only executed for diagonal conceptors. For conceptors, this was not necessary, however, there was a washout period required after the morphing period.
\doubleEnter
The first three dimensions of the $61$-dimensional output were plotted and are shown in Figures \ref{fig:c_hm_morphing} and \ref{fig:dc_hm_morphing} for conceptors and diagonal conceptors, respectively. The left and right columns in both figures show patterns $j_{1}$ (boxing) and $j_{2}$ (cartwheel). The middle column in both figures shows how pattern $j_{1}$ is morphed to pattern $j_{2}$. In the case of conceptors, it seems that the morph combines the characteristics of both patterns, where the beginning of the morph exhibits more characteristics of pattern $j_{1}$ and the end of the morph shows more characteristics of pattern $j_{2}$, as is expected. In the case of diagonal conceptors, the morphing period shows less characteristics of both patterns. This is due to the fact that the reservoir dynamics are nudged towards the correct starting state of pattern $j_{2}$, as described earlier.
\doubleEnter
At present, a more in-depth analysis of a morph with diagonal conceptors remains mostly unexplored. For example, it may be interesting to research how the geometric properties of the diagonal conceptors are carried through the morph. Therefore, any conclusions that are drawn are based on visual inspection rather than mathematical rigour. 
\doubleEnter
Diagonal conceptors are less suitable for morphing patterns than conceptors. In the case of the periodic patterns, they performed well, as the morphing period did not seem to contain any sudden changes in reservoir dynamics. However, this was not the case for the chaotic attractors. The self-generated pattern seemed to exhibit unpredictable behavior during the morphing period for diagonal conceptors, whereas for conceptors the self-generated pattern showed less sudden changes in the reservoir dynamics. The same was true for the human motions, where the reservoir required to be nudged during the morphing period in the case of diagonal conceptors.

\newpage
\section{Online adaptation}
\label{sec:online_adaptation}
\begin{figure}[ht]
    \centering
    \includegraphics[width=\textwidth]{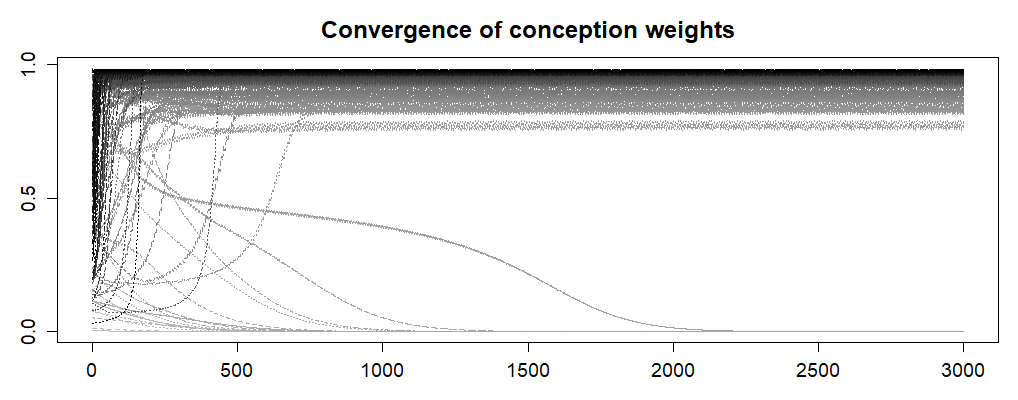}
    \caption{Convergence of the conception weights $c_{i}$ over the span of $n=3000$ time steps.}
    \label{fig:convergence_conception_weights}
\end{figure}

\begin{figure}[ht]
    \centering
    \includegraphics[width=\textwidth]{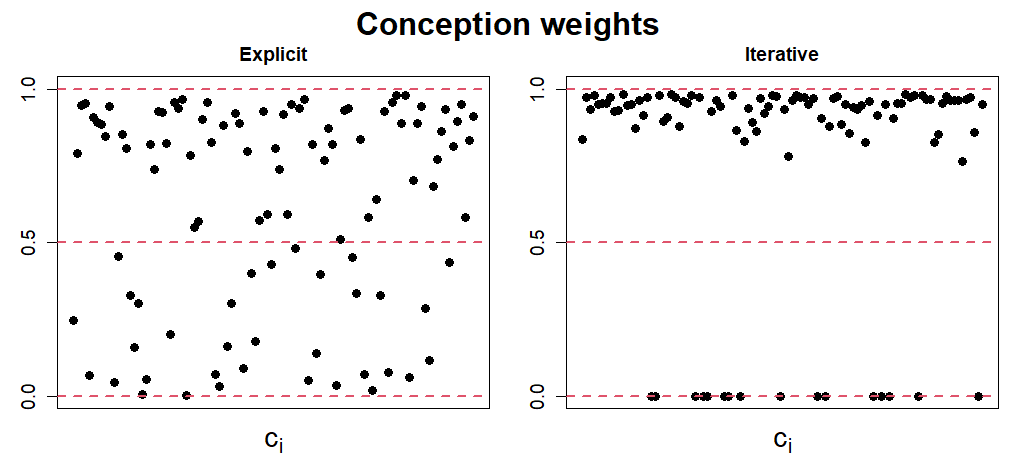}
    \caption{The conception weights after stage 1 for both the iterative and explicit simulation.}
    \label{fig:conception_weights}
\end{figure}

As mentioned in Section \ref{sec:diagonal_autoconceptors}, diagonal conceptors can be trained by online adaptation, similar to autoconceptors. The section dedicated to autoconceptors in the conceptors report comprises various examples and insights into the workings of these autoconceptors \cite{MonsterReport}. Diagonal autoconceptors were briefly introduced in Section \ref{sec:diagonal_autoconceptors} for the sake of completeness, but they are not analysed to the same extent as autoconceptors in the conceptors report. Nevertheless, this section demonstrates how diagonal autoconceptors could be implemented through an easy example using the online adaptation rule from Equation \ref{eq:diagonal_autoconceptors_adaptation_rule}.
\doubleEnter
The patterns in this example are the same as in Section \ref{sec:simulations_periodic_patterns}, initialized with all the corresponding parameters. For notational simplicity, the superscript $j$ is omitted. The simulation is conducted almost identically, except that during stage 1, the conception weights are adapted every time step according to the adaptation rule
\begin{equation}
c_{i}(n+1)=c_{i}(n)+\lambda_{i}\big((1-c_{i}(n))z_{i}^{2}(n)-\alpha^{-2}c_{i}(n)\big),
\end{equation}
where $c_{i}$ is the conception weight of neuron $i$, $\lambda_{i}$ is the learning rate associated with conception weight $i$, $\alpha$ is the aperture, and $z_{i}(n)$ is the $i$-th element of $\mathbf{z}(n)$ at time step $n$. The learning rate was set to $\lambda_{i}=0.5$ for all $i$ and the length of the adaptation period was set to $n_{stage1}=3000$ to ensure convergence. Note that true convergence would only be achieved in the limit of $\lambda_{i}\to0$. However, for the purpose of this simulation, true convergence is not necessary. The conception weights were collected over the whole stage 1 period and plotted as a function of time, which is shown in Figure \ref{fig:convergence_conception_weights}. It can be seen that the conception weights converge to either $0$ or a value in the range $[0.5,1]$, which confirms what was derived in Section \ref{sec:diagonal_autoconceptors}.
\doubleEnter
The simulation was also conducted without the adaptation rule, i.e., the diagonal conceptors were computed at the end of stage 1 according to Equation \ref{eq:ci_equation}, to allow for comparison. Computing the diagonal conceptors with an adaptation rule will be referred to as computing the diagonal conceptors \textit{iteratively} and without adaptation rule will be referred to as \textit{explicitly}. At this point it should be noted that both the iterative and explicit method yielded good accuracy for this example. The conception weights after stage 1 for both the iterative and explicit method simulation are shown in Figure \ref{fig:conception_weights}.
\doubleEnter
The explicitly computed conception weights are distributed over the range $[0,1]$, where it appears slightly denser towards $1$. Increasing the aperture will shift more conception weights towards $1$. The iteratively computed conception weights are either $0$ or in the range $[0.5,1]$, as discussed in Section \ref{sec:diagonal_autoconceptors}. Note that some conception weights are $0$, which means that some neurons will be completely suppressed. For this reason, it can be hypothesized that the iterative method fails for more complex problems. However, this remains unexplored.

\newpage
\section{Discussion} 
\label{sec:discussion}
Diagonal conceptors remain mostly unexplored at the present moment. However, this section dives a bit deeper into some of the features.
\doubleEnter
From the simulations in the previous section, it was clear that, for the most part, diagonal conceptors work just as well as conceptors. Consequently, diagonal conceptors offer an alternative, computationally efficient architecture for controlling a recurrent neural network. Aside from their advantages, diagonal conceptors also have their disadvantages. Both will be discussed in this section. 
\doubleEnter
This section will attempt to analyse the dynamics of diagonal conceptors and how they relate to, and are different than, conceptors. Some of the steps in the training scheme will be analyzed, specifically, the randomly initialized diagonal conceptors. Furthermore, the advantages and disadvantages will be considered as well as some guidelines on how to manually train diagonal conceptors.

\subsection{Training diagonal conceptors}
This section collects several characteristics of diagonal conceptors, all related to the training algorithm. First, the use of randomly initialized diagonal conceptors is analyzed. Second, the influence of the aperture and the regularization constants $\varrho^{W^{out}}$ and $\varrho^{W}$ is explored. Lastly, the necessary stage 1 of the training algorithm is discussed. 

\subsubsection{Initial Diagonal Conceptors}
\label{sec:discussion_initial_diagonal_conceptors}
\begin{figure}[t]
    \centering
    \includegraphics[width=\textwidth]{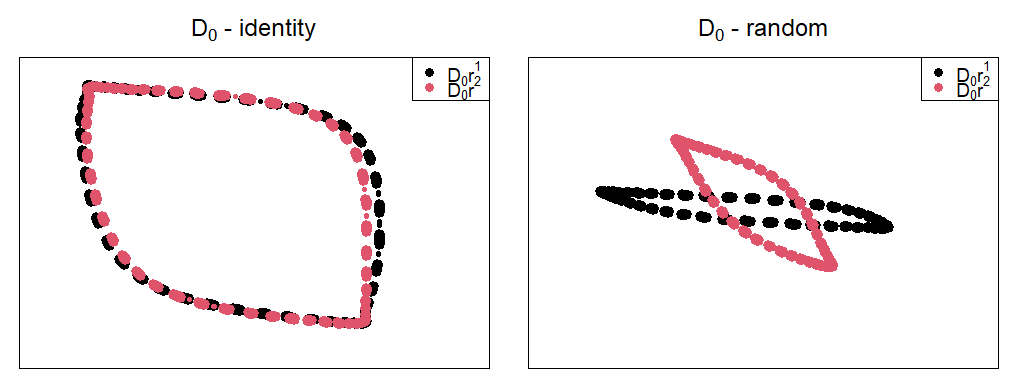}
    \caption{The neuron states during stage 1 of two neurons are depicted. In the left plot, the initial diagonal conceptors were initialized as identity. In the right plot, the initial diagonal conceptors were initialized randomly. The black dots show states representing pattern $1$. The red dots show states associated with pattern $2$.}
    \label{fig:initial_diagonal_conceptors}
\end{figure}

Let the reservoir comprise $N$ neurons. For simplicity, it is assumed that the neurons are non-leaky, i.e., the leaking rate is set to $\alpha_{l}=1$. The training of diagonal conceptors starts with initializing a number of parameters, specifically, and of interest, the initial diagonal conceptors $D^{j}_{0}=\text{diag}(\mathbf{c_{0}}^{j})$, where $\mathbf{c_{0}}^{j}=[c_{1}^{j}\;c_{2}^{j}\;...\;c_{N}^{j}]^{T}$ is the conception vector and $c_{i}^{j}$ are the conception weights, for all $i=1,2,...,N$. In contrast to the conceptors training scheme, randomly initialized diagonal conceptors are inserted in the update equations from the start, yielding 
\begin{equation}
\label{eq:dis_diag_update}
\begin{aligned}
\mathbf{r}(n+1)&=\tanh{(W^{*}\mathbf{z}(n)+W^{in}\mathbf{p}(n+1)+\mathbf{b})},\\
\mathbf{z}(n+1)&=D_{0}\mathbf{r}(n+1).
\end{aligned}
\end{equation}
The reservoir is run according to Equation \ref{eq:dis_diag_update} for an initial washout period of $n_{washout}$ time steps and then for another $n_{stage1}$ time steps. To understand the necessity of these initial diagonal conceptors in the loop, it is helpful to recap the idea behind conceptors.
\doubleEnter
If a reservoir is driven by pattern $j$, it creates a point cloud of reservoir states. The geometry of this point cloud is captured by the conceptor associated with pattern $j$. The conceptor can then be used to constrain the dynamics of the reservoir by projecting reservoir states onto the point cloud. A conceptor can distinguish similar patterns well, because their high degrees of freedom allows for minor adjustments in the projection onto the point cloud. However, diagonal conceptors have significantly fewer degrees of freedom and will have more trouble distinguishing similar patterns. This is where the initial diagonal conceptors come in. The initial diagonal conceptors create a random scaling of the reservoir states $\mathbf{r}(n)$. Intuitively, the initial diagonal conceptor shrinks the area in state space associated with pattern $j$ differently for each pattern $j$. Therefore, the reservoir states, $\mathbf{z}(n)=D_{0}\mathbf{r}(n)$, that are created are still characteristic for each pattern. Similar patterns, like the two sine waves in Section \ref{sec:simulations_periodic_patterns}, are "pulled apart", making it easier for the diagonal conceptors to determine which state to follow.
\doubleEnter
Interestingly, this initial random scaling strategy can be extended to matrix conceptors, which provides a novel way of working with conceptors. This is worth exploring further as it allows for a different training scheme for conceptors, which can be compared to the current training scheme described in Section \ref{sec:training_conceptors}. This comparison was not explored in this thesis. However, training conceptors with the random scaling was briefly attempted for the periodic patterns that are described in Section \ref{sec:simulations_periodic_patterns} and the simulation yielded good results, as defined in Section \ref{sec:simulation_methods_pattern_comparison}. The simulation can be found in the aforementioned GitHub repository \cite{GithubRepoDiagonalConceptors}.
\doubleEnter
Returning to diagonal conceptors, the two sine waves from the periodic patterns example from Section \ref{sec:simulations_periodic_patterns} are taken as an example to show in more detail how similar patterns are "pulled apart". The sine wave with period $\approx8.83$ is called pattern $1$ and the sine wave with period $\approx9.83$ is called pattern $2$. The simulation is conducted twice. The first time with the diagonal conceptors initialized to the identity matrix and the second time with randomly initialized diagonal conceptors. Regarding terminology it should be noted that initializing the diagonal conceptors to the identity matrix or the conception vector to the all-ones vector or initializing no diagonal conceptors all mean the same thing. In the first simulation, $c_{i}=1$ and in the second run $c_{i}$ is drawn from the uniform distribution on $[0,1]$, for all $i$. In both simulations, the reservoir states $\mathbf{z}^{j}=D^{0}\mathbf{r}^{j}$, for $j=1,2$, are collected during stage 1 of length $n_{stage1}=500$. The results show the neuron state of two selected neurons for both pattern $1$ and pattern $2$ in both cases. Figure \ref{fig:initial_diagonal_conceptors} depicts the findings.
\doubleEnter
It is evident from Figure \ref{fig:initial_diagonal_conceptors} that the neuron states for the randomly initialized diagonal conceptors are more distinguishable than the neuron states for the identity initialized diagonal conceptors. In the left plot it is seen that the neuron states of the sine waves are very similar, hence difficult for the diagonal conceptors to tell apart. In the right plot, the scaled neuron states have been shrunk randomly and are now much easier to tell apart by the diagonal conceptors.
\doubleEnter
A few remarks can be made here. First, the randomness of the initialized diagonal conceptors is of a big influence on the training scheme. In general it is good practice to set a seed for computer simulations for reproducibility. However, in the case of diagonal conceptors this is vital, because underlying the training of diagonal conceptors lies the random initialization of the initial diagonal conceptors. Therefore, changing the seed of the simulation will most likely break the simulation. Second, there probably exists an optimal initialization. In this report, the conception weights were drawn from a uniform distribution on $[0,1]$, but this does not have to be optimal. One could experiment with other distributions and other ranges. A few experiments were conducted where the conception weights were drawn from the uniform distribution on $[-1,1]$ and the standard normal distribution. Both yielded unsatisfactory results. In the case of the former, this could be attributed to the fact that negative conception weights change the sign of negative neuron states, which could disrupt the reservoir dynamics. In the case of the latter, it is not guaranteed that a neuron state stays in the range $[-1,1]$, which can also disrupt the reservoir dynamics. However, neither cases were analyzed thoroughly.

\subsubsection{Aperture and Regularization}
\begin{figure}[ht]
    \centering
    \includegraphics[width=\textwidth]{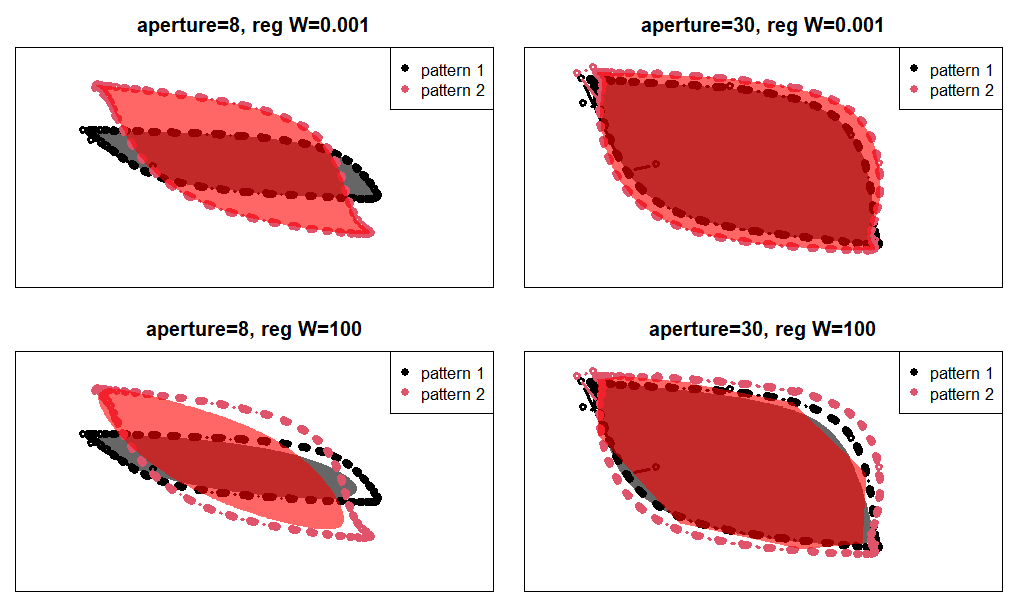}
    \caption{The neuron states during stage 2 (dots) and the neuron states during the self-generation period (filled-in areas) of two neurons are shown. The black color represents the neuron states associated with pattern $1$ and the red color represent the neuron states associated with pattern $2$. For explanation see text.}
    \label{fig:aperture_regularization}
\end{figure}

Besides the randomly initialized diagonal conceptors, the aperture $\alpha$ and regularization constant for recomputing the reservoir weights $\varrho^{W}$ play a big role in training the diagonal conceptors. Regarding the regularization constant for computing the output weights, $\varrho^{W^{out}}$, it was found that in most cases it was of little influence on the output. Therefore, it will not be discussed here. To demonstrate how $\alpha$ and $\varrho^{W}$ are of influence on the reservoir dynamics, the example from the previous section is continued. Four simulations were conducted where $\alpha$ and $\varrho^{W}$ were changed. The results are shown in Figure \ref{fig:aperture_regularization} and require a bit of explaining.
\doubleEnter
The aperture was either set to $\alpha^{j}=8$ or $\alpha^{j}=30$, for $j=1,2$, and $\varrho^{W}$ was either to $\varrho^{W}=0.001$ or $\varrho^{W}=100$. The reservoir was run with randomly initialized diagonal conceptors for $n_{stage1}=500$ time steps, after which the diagonal conceptors $D^{j}$ were computed. This computation depends heavily on $\alpha^{j}$. Afterwards, the reservoir was continued to be driven for $n_{stage2}=500$ time steps, during which the states $\mathbf{z}^{j}=D^{j}\mathbf{r}^{j}$ were collected, for $j=1,2$. Then, the patterns were stored in the reservoir, which is regulated by $\varrho^{W}$. Finally, the reservoir was run for $n_{run}=500$ time steps with the newly computed reservoir weights. During this period, the states $\mathbf{z}^{j}=D^{j}\mathbf{r}^{j}$ were again collected. 
\doubleEnter
Similar to Figure \ref{fig:initial_diagonal_conceptors}, two neuron states where shown, yielding a two-dimensional plot for each simulation. The black color represents the neuron states associated with pattern $1$ and the red color represent the neuron states associated with pattern $2$. The dotted patterns are the states collected during stage 2 and the filled-in areas represent the states collected during the self-generation period, where the states lie on the boundary of the area. The area within this boundary is filled in for aesthetic ease.
\doubleEnter
From Figure \ref{fig:aperture_regularization}, a few observations can be made. First, increasing the aperture decreases the differences between the states. Consequently, for a large aperture, the point cloud depicted in the left left plot of Figure \ref{fig:initial_diagonal_conceptors}, where $D_{0}=I$, is recovered. The states in the left column of Figure \ref{fig:aperture_regularization} seem to be a combination of the right and left plot of Figure \ref{fig:initial_diagonal_conceptors}. By adjusting the aperture it was found that the shape of the states, as seen in Figure \ref{fig:aperture_regularization}, tends towards the left and right plot in Figure \ref{fig:initial_diagonal_conceptors} for large and small apertures, respectively. Therefore, increasing the aperture appears to remove the advantage of the randomly initialized diagonal conceptors. This was also seen in the chosen apertures for the diagonal conceptors in the simulations in Section \ref{sec:simulations}, which were consistently smaller than the apertures for the conceptors. Second, the regularization constant $\varrho^{W}$ behaves like a regularization constant is expected to behave. In ridge regression, the regularization constant mitigates between the bias and the variance, where an increase in the regularization constant leads to an increase in the bias, but a decrease in the variance. Therefore, a large $\varrho^{W}$ will compress the shape of the states, thus decreasing the variance. What this means in practice is that if the aperture is increased it will be likely that $\varrho^{W}$ should be decreased. This phenomenon can be seen most clearly in the right column of Figure \ref{fig:aperture_regularization}. As the aperture is increased, the states will become more similar. However, if $\varrho^{W}$ is small, it will capture the nuances that are distinct for each pattern. If $\varrho^{W}$ is large, those nuances are suppressed, leading to indistinguishable state patterns. This is seen clearly in the bottom right plot of Figure \ref{fig:aperture_regularization}, where the gray and red areas almost completely overlap.

\subsubsection{Stage 1}
\label{sec:stage1}
Diagonal conceptors require a period during which the diagonal conceptors are computed, stage 1, as described in Section \ref{sec:training_diagonal_conceptors}. The states in that period can not be used for storing the patterns in the reservoir, which means that no information of that period is taken into account in stage 2. This information is lost, like in the washout period. Therefore, the learned pattern will be shorter than for conceptors. This was seen in the human motion simulation. The recalled patterns were shorter for diagonal conceptors compared to conceptors. In the case of the periodic patterns or the chaotic attractors, this was not a problem, because the patterns were generated synthetically.
\doubleEnter
As mentioned in the human motion simulations, in Section \ref{sec:simulations_human_motion}, a solution could be to first run stage 1 to compute the diagonal conceptors. Then, the pattern from stage 1 would be reused in stage 2, yielding much more training data for storing the patterns. This has successfully been attempted for the periodic patterns, chaotic attractors, and human motions, in simulations that can be found in the aforementioned GitHub repository \cite{GithubRepoDiagonalConceptors}. In all three cases, some of the parameters had to be adjusted to yield results with a good accuracy. For example, the aperture in the periodic patterns simulation needed to be decreased from $8$ to $7$, and in the chaotic attractors simulation, the aperture of pattern $2$ needed to be increased from $6$ to $7$. After these adjustments, the periodic patterns and chaotic attractors performed equally well as in the simulations described in Sections \ref{sec:simulations_periodic_patterns} and \ref{sec:simulations_chaotic_attractors}. In the case of the human motions, the number of reservoir neurons had to be increased from $N=1000$ to $N=1500$ in order to yield results similar to the simulations in Section \ref{sec:simulations_human_motion}. From these simulations it can be concluded that reusing the pattern from stage 1 in stage 2 should be considered in future simulations of diagonal conceptors. In the case of the periodic patterns or the chaotic attractors it was not necessary, but in the case of the human motions it made a significant improvement, as the learned patterns were notable longer. Therefore, for a set of patterns such as the human motions, where patterns may be transient or irregular stochastic it is advised to apply this adjustment in the training scheme.

\subsection{Stability}
\label{sec:discussion_stability}
This section starts with the following observation. For a reservoir comprised of $N$ neurons, a conceptor matrix $C$ has $(N+1)N/2$ degrees of freedom\footnote{A conceptor matrix is symmetric, so it has $N+(N-1)+(N-2)+...+1=N(N+1)/2$ degrees of freedom.}. A diagonal conceptor $D$ has only $N$ degrees of freedom. Therefore, as the number of reservoir neurons increase, the degrees of freedom of a conceptor matrix scales much faster than the degrees of freedom of a diagonal conceptor matrix. Because of this nonlinear scaling, one expects that the number of reservoir neurons should be significantly larger in the diagonal conceptors simulations compared to the conceptors simulations. However, the simulations in Section \ref{sec:simulations} showed that this was not necessary. In the case of the periodic patterns and the chaotic attractors, the number of reservoir neurons was set equal for conceptors and diagonal conceptors. Even in the case of the human motions, which comprised many more dimensions ($61$) compared to the periodic patterns ($1$) and the chaotic attractors ($2$), the number of reservoir neurons for diagonal conceptors was only $1000/600=1\frac{2}{3}$ times larger than for conceptors. The fact that the number of reservoir neurons for diagonal conceptors does not scale faster is mainly due to the different dynamics of diagonal conceptors imposed by the randomly initialized diagonal conceptors. Nevertheless, it must be noted that diagonal conceptors show slightly more unstable behavior than conceptors.
\doubleEnter
In the periodic patterns and chaotic attractors simulations, the diagonal conceptors were robust against perturbations. In the case of the chaotic attractors, this was especially evident in the morphing simulation, where the self-generated pattern was unpredictable during the morphing period, but afterwards it stabilized again to the desired pattern. Furthermore, self-generation with diagonal conceptors was usually initiated from the all-zeros vector, but other starting states were tried. It was found that any starting state in $\mathbb{R}^{N}$ yielded stable patterns. This stable behavior was found for both the periodic patterns and the chaotic attractors. Regardless of the starting state, the correct pattern was always engaged in.
\doubleEnter
In addition, the parameters of the periodic patterns simulation were quite robust. Large variations in the parameter value still preserved stability. For the chaotic attractors simulation, the parameters were less robust, as a variation in the scaling of the weights could disrupt the stability of the self-generated patterns. Nevertheless, the parameters could still be varied to a certain extent, while stability of the self-generated patterns was retained. In the case of the human motion simulation, setting the parameters required a bit more patience. It was shown that the found parameters could not be varied much, as it would disrupt the stability of the most complex patterns. Overall, the stability of diagonal conceptors depends on the complexity of the task, but diagonal conceptors are more stable than one would initially expect.

\newpage
\section{Conclusion} 
\label{sec:conclusion}
Throughout this report, diagonal conceptors have been shown to be a practical alternative to conceptors. Surprisingly, what is gained in computation efficiency is not immediately lost in stability. Nevertheless, diagonal conceptors are more unstable than conceptors for more complex tasks.
\doubleEnter
In the simulation of the periodic patterns, in Section \ref{sec:simulations_periodic_patterns}, the only deficit of diagonal conceptors was the ability to extrapolate the periods of two sine waves in the morphing. However, besides that, diagonal conceptors perform just as well as conceptors. The fact that diagonal conceptors can distinguish between similar sine wave patterns is mainly due to the randomly initialized diagonal conceptors. They are inserted in the update equations from the beginning to randomly scale the reservoir states. This training is different from conceptors, but proved to be vital for the performance of diagonal conceptors.
\doubleEnter
In the simulation of the chaotic attractors, in Section \ref{sec:simulations_chaotic_attractors}, diagonal conceptors show that they are also capable of capturing the dynamics of more complicated systems. Furthermore, they proved to be stable, as perturbations in the reservoir dynamics would quickly be resolved. This could be seen by the fact that they would yield stable self-generated patterns from any starting state in $\mathbb{R}^{N}$. Furthermore, the patterns could be morphed, but during the morphing period the pattern showed unpredictable behavior. Nevertheless, after the morphing period, the diagonal conceptors were able to stabilize the self-generated pattern to the desired pattern. Furthermore, in this simulation it was shown that the aperture parameter for diagonal conceptors does not offer the same amount of control as it does for conceptors. An adjustment of the aperture of a single pattern can disrupt the stability of other patterns. This is due to the fact that the diagonal conceptors are computed before the patterns are stored in the reservoir, which creates a delicate balance between the aperture and storing the patterns in the reservoir. This balance is mitigated between the aperture and the regularization constant for recomputing the reservoir weights. Conceptors, on the other hand, are computed independently of storing the patterns, hence the aperture offers much more control over the stability of the self-generated patterns. 
\doubleEnter
In the simulation of the human motions, in Section \ref{sec:simulations_human_motion}, diagonal conceptors were able to recall the learned patterns. After the learned patterns had been recalled, the self-generated continuation departed from the kind of motion that the initial training data represented. Note that this is expected for transient patterns, but not for periodic patterns. The diagonal conceptors were not "strong" enough to guide unknown states back to a learned state. stage 1 shortens the length of the training data, but an alternative approach was proposed in Section \ref{sec:stage1}, where the pattern in stage 1 is reused in stage 2.

\newpage
\section{Future Work}
\label{sec:future_work}
The research of diagonal conceptors is still in the early stages. Some features were already discussed in Section \ref{sec:discussion}, but there is much potential for future work, for example:
\begin{itemize}
    \item Optimizing the randomly initialized diagonal conceptors.
    \item Researching diagonal autoconceptors, similar to the autoconceptors that are discussed in Section 3.14 of the conceptors report \cite{MonsterReport}.
    \item Improving the delicate balance between the aperture and the regularization constants for storing the pattern in the reservoir.
    \item Researching how and if the other features of conceptors translate to diagonal conceptors. They features include, but are not limited to: Boolean operations, neural memory management, recognizing patterns.  
\end{itemize}
The last point is of the most importance, as the comprehensive list of capabilities of conceptors is long and a practical implementation of those features could open up further research in many fields.

\end{document}